\documentclass[10pt,twocolumn,letterpaper]{article}

\usepackage{iccv}              
\usepackage{graphicx}
\usepackage{amsmath}
\usepackage{graphicx}    
\usepackage{float}       
\usepackage{xcolor}      
\usepackage{amssymb}     
\usepackage{wrapfig}     
\usepackage{multirow}    
\usepackage{multicol}    
\usepackage{natbib} 
\usepackage{enumitem}
\usepackage{amsmath}

\definecolor{iccvblue}{rgb}{0.21,0.49,0.74}
\usepackage[pagebackref,breaklinks,colorlinks,allcolors=iccvblue]{hyperref}

\def\paperID{5176} 
\def\confName{ICCV}
\def\confYear{2025}

\newcommand{\equalcontrib}{\textsuperscript{*}}

\title{DATA: Domain-And-Time Alignment for High-Quality Feature Fusion in Collaborative Perception}

\author{
Chengchang Tian\textsuperscript{1}\equalcontrib, 
Jianwei Ma\textsuperscript{1}\equalcontrib, 
Yan Huang\textsuperscript{1$\dagger$}, 
Zhanye Chen\textsuperscript{1$\dagger$}, 
Honghao Wei\textsuperscript{2},
Hui Zhang\textsuperscript{1}, 
Wei Hong\textsuperscript{1}\\
\textsuperscript{1} Southeast University, Nanjing, China\\
\textsuperscript{2} Washington State University, Pullman, WA, USA\\
{\tt\small \{chengchang\_tian, jianwei\_ma, yan\_huang, chenzhanye, huizhang, weihong\}@seu.edu.cn}\\
{\tt\small honghao.wei@wsu.edu}
}
\begin{document}
\allowdisplaybreaks
\maketitle
\footnotetext[1]{$^*$Equal contribution}
\footnotetext[2]{$^\dagger$Corresponding authors}
\begin{abstract}
\noindent Feature-level fusion shows promise in collaborative perception (CP) through balanced performance and communication bandwidth trade-off. However, its effectiveness critically relies on input feature quality. The acquisition of high-quality features faces domain gaps from hardware diversity and deployment conditions, alongside temporal misalignment from transmission delays. These challenges degrade feature quality with cumulative effects throughout the collaborative network. In this paper, we present the \textbf{D}omain-\textbf{A}nd-\textbf{T}ime \textbf{A}lignment (DATA) network, designed to systematically align features while maximizing their semantic representations for fusion. Specifically, we propose a Consistency-preserving Domain Alignment Module (CDAM) that reduces domain gaps through proximal-region hierarchical downsampling and observability-constrained discriminator. We further propose a Progressive Temporal Alignment Module (PTAM) to handle transmission delays via multi-scale motion modeling and two-stage compensation. Building upon the aligned features, an Instance-focused Feature Aggregation Module (IFAM) is developed to enhance semantic representations. Extensive experiments demonstrate that DATA achieves state-of-the-art performance on three typical datasets, maintaining robustness with severe communication delays and pose errors. The code will be released at \url{https://github.com/ChengchangTian/DATA}.
\end{abstract}
\section{Introduction}
\label{sec:intro}

Collaborative perception (CP) \cite{cp1, cp2, cp3} has emerged as a crucial solution to overcome the inherent limitations of single-agent perception \cite{v2vnet, opv2v}, such as limited perception range and occluded areas. By enabling multiple agents to share their own perceptual information, CP facilitates a more comprehensive understanding of surroundings. 
\begin{figure}[!t]
    \centering
    \hfill  
    \begin{minipage}{0.5\textwidth}  
        \centering  
        \includegraphics[width=0.9\textwidth]{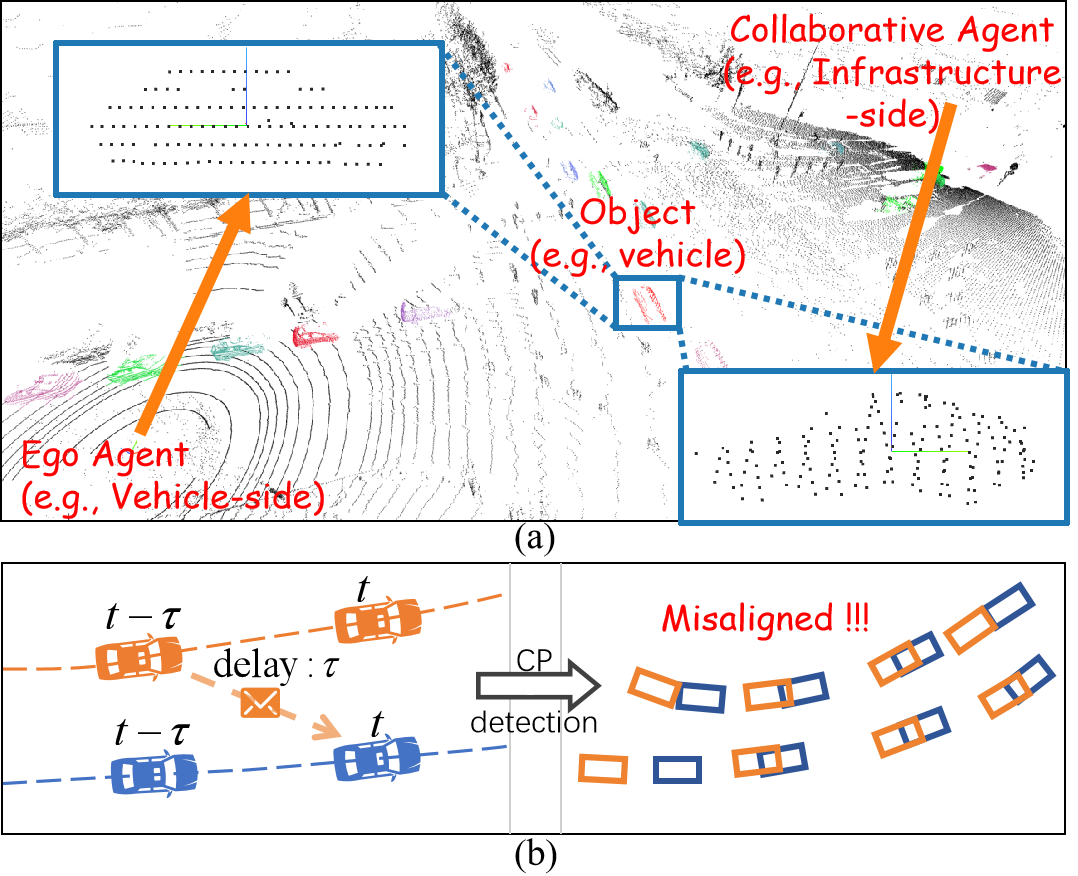}
        \vspace{-0.3cm}
        \caption{(a) Domain gap. E.g., different structured and scattered foreground patterns of different agents. 
        (b) Temporal Misalignment. E.g., communication latency in the collaboration process and the followed observation misalignment between agents.}
        \label{fig:motivation}
    \end{minipage}
    \vspace{-0.8cm}
\end{figure}

For CP, intermediate fusion \cite{coca3d, disconet, when2com, who2com, umc, v2xvit, how2comm}, which operates at the feature level for information sharing and integration, has been extensively studied due to its balanced trade-off between perception performance and communication bandwidth. However, it is difficult to maintain high-quality input features for fusion. In real-world deployments, obtaining high-quality features during the feature acquisition phase is fraught with significant challenges \cite{v2v4real}. Hardware heterogeneity \cite{v2xvit, div2x, dairv2x, v2xsim} (\eg, LiDARs with varying numbers of laser beams and diverse modes for acquiring point clouds) and differences in agent conditions (\eg, sensor mounting heights and angles) result in distinct data distributions among agents. This divergence in data among agents creates a domain gap in CP, as illustrated in Figure \ref{fig:motivation}(a). Furthermore, temporal delays \cite{v2vnet, syncnet, ffnet, cobevflow} are introduced during communication transmission between agents, as illustrated in Figure \ref{fig:motivation}(b), causing the features of the same object to be misaligned. The domain and time misalignment not only significantly degrades the quality and reliability of the acquired features, but also exhibits cumulative effects throughout the collaborative system. Hence, domain and time alignment during feature acquisition is crucial to the precision and robustness of CP.

Various attempts have been made to achieve domain and time alignment during feature acquisition. For domain alignment, DI-V2X \cite{div2x} achieves domain-invariant representations through a reference domain constructed by randomly mixing instances from different sources. However, this mixing approach compromises physical validity by disrupting occlusion relationships between objects in the reference domain. For temporal alignment, the methods based on global motion flow, like FFNet \cite{ffnet}, model the entire scene to construct future frame features.  However, their scene-wide optimization is biased by dominant background regions, undermining the modeling of fine-grained foreground motion patterns. In contrast, RoI-based methods, like CoBEVFlow \cite{cobevflow}, employ localized motion prediction for focused modeling. However, their reliance on region proposals impairs occlusion handling and recall, limiting the understanding of scene-wide motion dynamics.

\begin{figure*}[!t]
    \centering
    \includegraphics[width=1\textwidth]{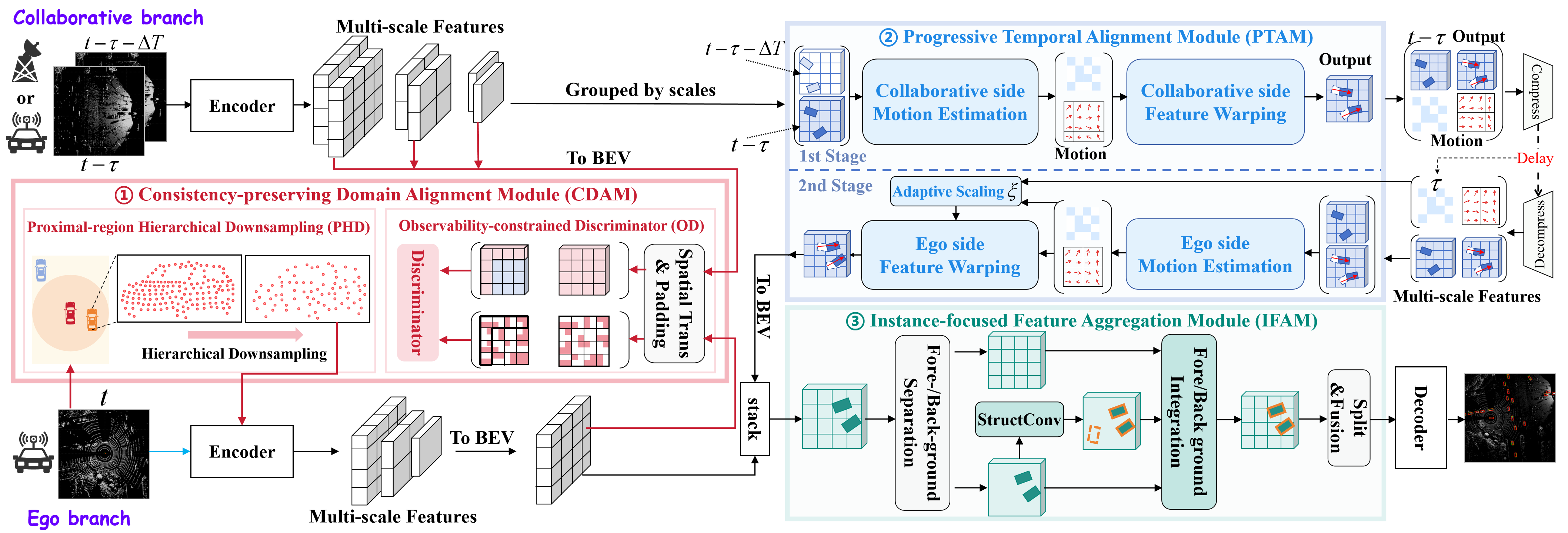}
    \vspace{-0.6cm}
    \caption{Overview of DATA. Arrows indicate data flow: black arrows represent streams used for both training and inference, {\color{red}red arrows} are used only during training, and {\color{blue}blue arrow} is used only during inference. Notation table in supplementary materials aids understanding.}
    \label{fig:overall}
    \vspace{-0.6cm}
\end{figure*}
To improve CP performance, we propose the \textbf{D}omain-\textbf{A}nd-\textbf{T}ime \textbf{A}lignment (DATA) network, generating high-quality features for 3D object detection. Specifically, it focuses on learning domain-invariant and time-coherent representation for robust feature acquisition. 
The DATA network consists of three main modules: (i) A Consistency-preserving Domain Alignment Module (CDAM) to reduce domain gaps in the training stage. The domain gaps, within a single agent and between agents,  are minimized through two complementary approaches: achieving the consistency of distance-adaptive point density while preserving physical validity via proximal-region hierarchical downsampling, and mitigating genuine domain gaps through feature-level adversarial learning under consistent observation conditions across shared regions; (ii) A Progressive Temporal Alignment Module (PTAM) to address temporal misalignment. It hierarchically captures motion patterns using multi-scale features and models complex motion through two-stage compensation to achieve scene-wide representation, and the multi-window self-supervised training strategy simultaneously enables effective foreground object motion learning and maintains global scene coherence; (iii) An Instance-focused Feature Aggregation Module (IFAM) to effectively aggregate the aligned features from multiple agents. To validate the effectiveness of DATA, we conduct extensive experiments on three typical CP datasets: DAIR-V2X-C \cite{dairv2x}, V2XSET \cite{v2xvit}, and V2XSIM \cite{v2xsim}. Comprehensive experimental results demonstrate that our method outperforms existing state-of-the-art methods by 2.36\% AP$_{70}$ on DAIR-V2X-C, and by 1.84\% and 2.85\% AP$_{70}$ on V2XSIM and V2XSET. DATA also exhibits exceptional delay robustness, maintaining 75.58\% AP$_{50}$ under 500ms communication delay, surpassing SOTA methods by 2.61\%. The main contributions of this paper can be summarized as follows:
\begin{itemize}
    \item We propose DATA, which is a new CP framework that primarily addresses challenges of feature acquisition through domain and time alignment, complemented by instance-level feature refinement to maximize the semantic representations of aligned features. 
    \item We design CDAM, PTAM, and IFAM to reduce domain gaps, achieve temporal feature coherence, and sufficiently exploit the semantic representations of aligned features. 
\end{itemize}

\section{Related Work}
\label{sec:relatedwork}

\noindent {\textbf{Object Detection in Collaborative Perception.}} 
Collaborative Perception is a crucial section in autonomous driving systems. Recently, various studies \cite{stamp, freealign, bevheight, bevheight++, transiff} have explored diverse approaches to improve perception performance. Where2comm \cite{where2comm} selectively transmits perceptually critical features via confidence maps. CodeFilling \cite{codefilling} further compresses features based on codebook-based encoding to reduce transmission cost. HM-ViT \cite{hmvit} proposes heterogeneous 3D graph transformers to handle varying sensor configurations between agents. HEAL \cite{heal} introduces a backward alignment training mechanism to construct a unified feature space. MRCNet \cite{mrcnet} tackles pose noise, perception noise, and motion blur through a motion-aware robust communication framework.

\noindent {\textbf{Domain Alignment in Collaborative Perception.}} Addressing the domain gap issue \cite{v2v4real} is a crucial step in enhancing collaborative perception performance. Recent approaches address domain gaps through different mechanisms. V2X-ViT \cite{v2xvit} addresses domain gaps by encoding different combinations of agents through specialized embeddings in its heterogeneous multi-agent self-attention module. MPDA \cite{MPDA} tackles domain gaps through a learnable feature resizer and sparse cross-domain transformer. DI-V2X \cite{div2x} proposes a distillation framework with domain-mixing instance augmentation and progressive distillation. In this paper, we propose density-aware and region-aware training mechanisms to better bridge the domain gap between various heterogeneous agents.

\noindent {\textbf{Time Alignment in Collaborative Perception.}} Temporal synchronization is also instrumental in determining real-world deployment performance. Initial efforts to address this challenge include V2VNet \cite{v2vnet} and V2X-ViT \cite{v2xvit}, which pioneer neural network approaches using convolutional networks and delay-aware positional encoding for delay compensation. Furthermore, SyncNet \cite{syncnet} extends these approach by incorporating multiple historical frames through a pyramid LSTM architecture. FFNet \cite{ffnet} introduces a flow-based framework to predict future features for aligned fusion. Meanwhile, CoBEVFlow \cite{cobevflow} combines RoI-based matching with transformer-based methods for motion prediction. In this paper, we propose two-stage compensation methods to capture both fine-grained and global motion patterns. In Additional, we introduce a multi-window self-supervised strategy to better learn the motion patterns of each object within local regions.

\section{Method}
\label{sec:method}

\subsection{Problem Formulation and Overall Architecture}

Our framework operates in a CP system with $N$ agents, where each agent simultaneously functions as both a data receiver and a data provider. 
In the whole paper, we define the agents, which are currently receiving and transmitting data, as the ego agent and collaborative agents. 
And subscripts $i$ and $j$ denote the ego agent and collaborative agents, where $i \neq j$. At the current time $t$, the ego agent processes its latest data ${\cal{X}}_i(t)$, while integrating data transmitted by the collaborative agents at timestamp $t-\tau$, where $\tau$ represents the transmission delay. To compensate for this transmission delay, the collaborative agent processes its two latest frames ${\cal{X}}_j(t-\tau)$ and ${\cal{X}}_j(t-\tau-\Delta T)$ before transmission, providing both perception and motion information.

The overall architecture of DATA is shown in Figure \ref{fig:overall}. \\
{\textit{(i) CDAM to Align Domain (Only in Training):}}  First, the point clouds of ego agent undergoes PHD of CDAM ({\textit{red part}}) to generate multi-scale features. In collaborative branch, it processes point clouds to generate multi-scale features for two timestamps. The features of both agents are converted to BEV features, then inputted to the OD of CDAM to facilitate domain alignment between both agents.\\
{\textit{(ii) PTAM to Align Time (in Training \& Testing):}} The multi-scale features of collaborative agent pass through the first stage of PTAM ({\textit{blue part}}). Then the output feature, along with the feature from the latest frame ($t-\tau$) and motion information, are compressed and transmitted to the ego agent. The ego agent decompresses the received data to recover the multi-scale features and implements the second stage of PTAM to further adjust the features based on the transmission delay $\tau$, achieving complete temporal alignment. \\
{\textit{(iii) IFAM to Fuse Features (in Training \& Testing):}} The temporally aligned multi-scale features are converted to BEV features, subsequently fed into IFAM ({\textit{green part}}) together with the ego's BEV features to fuse the complementary information into a comprehensive representation. This fused representation is served as input for the detection head (decoder) to produce the final detection results.
\subsection{Consistency-preserving Domain Alignment Module (CDAM)}\label{CDAM}
Domain gaps from hardware heterogeneity and deployment variations manifest at raw data and feature levels.  Therefore, we introduce CDAM to address these through PHD for raw data processing and OD for feature-level alignment.

\subsubsection{Proximal-region Hierarchical Downsampling (PHD)\label{PHD}} Point clouds of a single agent exhibit significant density variations with varying observation distances \cite{v2v4real}, biasing the model learning toward high-density regions and affecting the feature extraction. To address this issue, PHD is proposed to balance the density distribution of point clouds at the ego agent. 
PHD mainly consists of three steps. 

\begin{figure}[!t]
    \centering
    \includegraphics[width=0.95\columnwidth]{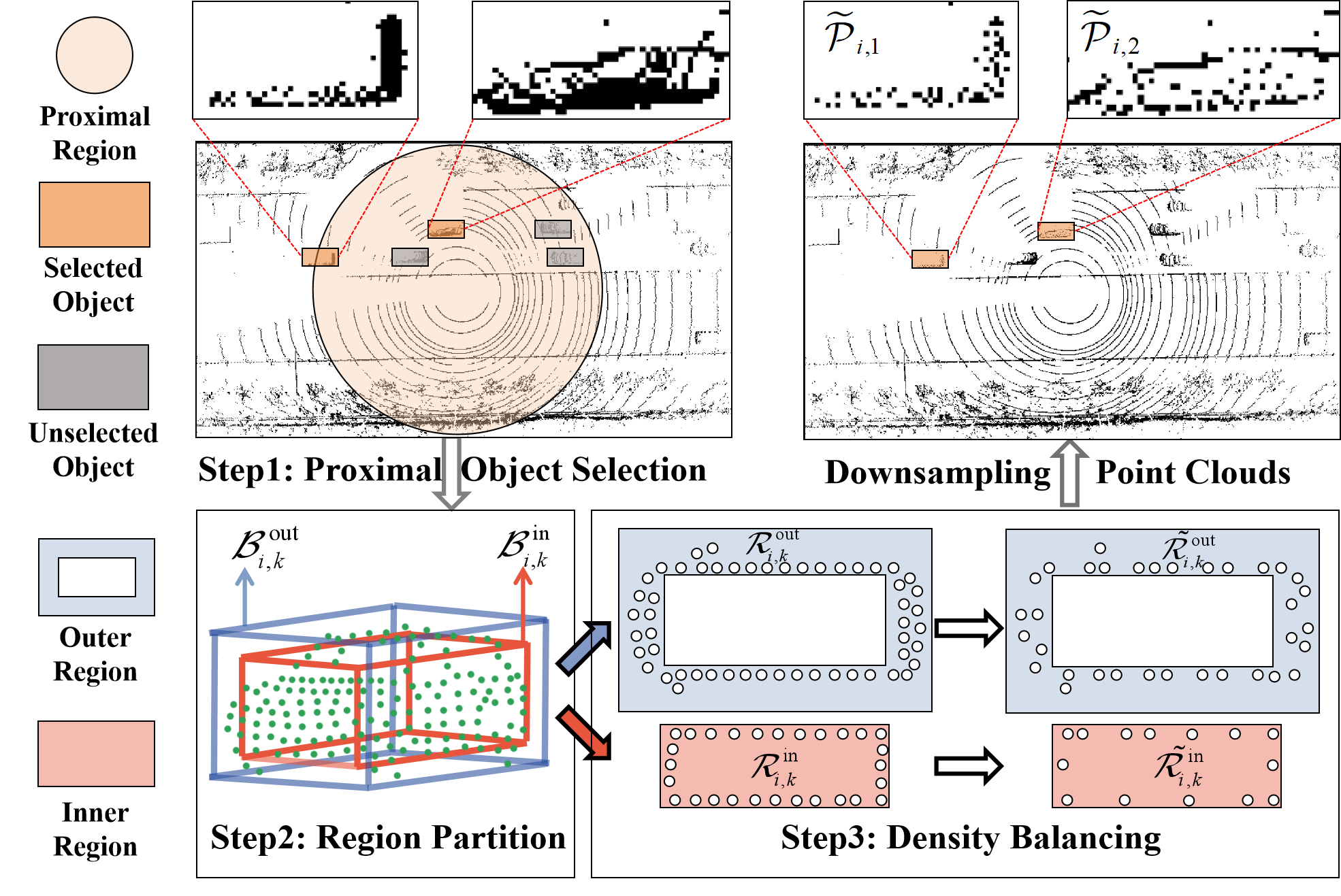}
    \vspace{-0.3cm}
    \caption{Pipeline of PHD. Point clouds processed by PHD simultaneously deliver contour preservation and density reduction.}
    \label{PipelineofPHD}
    \vspace{-0.5cm}
\end{figure}
{\textbf{Step 1: Proximal Object Selection.}} For the input scene, we define the set of all objects observable to ego agent $i$ as ${\cal{O}}_i=\{o_1, o_2, ..., o_{N^\text{total}_i}\}$, where $N^\text{total}_i$ is the total number of the observable objects. Objects within the proximal region are identified based on their distance to the ego agent as
\begin{equation}
{\cal{O}}^\text{prox}_i = \{\;o_k\;|\;d_i(o_k) \leq d_{th}, k=1,\dots,N^\text{total}_i\},
\label{eq:rangesel}
\end{equation}
\noindent where $d_i(o_k)$ is the distance from the ego agent $i$ to the $k$-th object and $d_{th}$ is the distance threshold. To maintain the distribution similarity with the original scene after downsampling, we select $N_\text{proc}$ objects from ${\cal{O}}^\text{prox}_i$ for subsequent processing. If $|{\cal{O}}^\text{prox}_i| \textgreater N_\mathrm{max}$, then $N_\text{proc}=N_\mathrm{max}$ objects are randomly selected from ${\cal{O}}^\text{prox}_i$, otherwise, all objects in ${\cal{O}}^\text{prox}_i$ are selected. Herein, $|\cdot|$ denotes the cardinality of a set and $N_\mathrm{max}$ is the predefined maximum number of objects.

{\textbf{Step 2: Region Partition.}} For each selected object, its point clouds are partitioned  into inner and outer regions by using concentric bounding boxes with different scales. Let ${\cal{B}}_{i,k}^\text{out}$ denote the oriented bounding box of the $k$-th object parameterized by center coordinates $(x_k, y_k, z_k)$, dimensions $(h_k, w_k, l_k)$, and orientation $\theta_k$.
We also define a scaling factor $\alpha \in (0,1)$ to adjust the height, width, and length of the bounding box. This creates an inner bounding box with the same center and orientation as ${\cal{B}}_{i,k}^\text{out}$, \ie, ${\cal{B}}_{i,k}^\text{in} = \{(x_k, y_k, z_k, \alpha h_k, \alpha w_k, \alpha l_k, \theta_k)\}$, where $k=1,\dots,N_\text{proc}$.
Then the point sets in the inner and outer regions are 
\begin{equation}
    {\cal{R}}_{i,k}^\text{in} = \{p | p \in {\cal{B}}_{i,k}^\text{in}\},\;\;{\cal{R}}_{i,k}^\text{out} = \{p | p \in {\cal{B}}_{i,k}^\text{out} \setminus {\cal{B}}_{i,k}^\text{in}\}.
\label{inregion}
\end{equation}
This partition effectively separates the sparse interior points from the dense points on the object contour.

\textbf{Step 3: Density Balancing.} Next, the Farthest Point Sampling (FPS) \cite{FPS} is applied with different ratios to the partitioned regions. For the inner region, a high downsampling ratio $\beta_\text{in}$ is used while a conservative downsampling ratio $\beta_\text{out}$ is applied to the outer region with more points left, thus capturing the object contour and preserving its geometric details. The downsampling can be formulated as
\begin{equation}
        \tilde{{\cal{R}}}_{i,k}^\text{in} = \text{FPS}({\cal{R}}_{i,k}^\text{in}, \beta_\text{in}),\;\;\tilde{{\cal{R}}}_{i,k}^\text{out} = \text{FPS}({\cal{R}}_{i,k}^\text{out}, \beta_\text{out}).
\label{insample}
\end{equation}
Finally, the point clouds of the $k$-th object, combining both inner and outer regions, can be formulated as $\tilde{{\cal{P}}}_{i,k} = \tilde{{\cal{R}}}^\text{in}_{i,k} \cup \tilde{{\cal{R}}}^\text{out}_{i,k},k=1,\dots,N_\text{proc}$.
Through this hierarchical downsampling, PHD enhances density-consistent representations for varying distances and preserves critical information of the original data, particularly occlusion relationships and geometric structures. 
\subsubsection{Observability-constrained Discriminator (OD)\label{OD}} In CP, the domain gaps between agents arise from both sensor-intrinsic properties and observation characteristics. To align these domain gaps, it is crucial to identify the regions where both agents maintain valid observations. 
To address this issue, we propose an OD module to explicitly incorporate observability into the domain alignment process.

First, we define the observations of ego agent $i$ as the ego domain, while the observations of $N-1$ collaborative agents are the collaborative domain. For domain discrimination, one collaborative agent is randomly selected for domain alignment. This enables diverse agent combinations that facilitate domain-invariant feature learning. 

Then, through a foreground estimator $\Phi(\cdot)$, the OD uses Bird's Eye View (BEV) features $H_i$ and $H_j$ of ego and collaborative agents to generate observability maps $M_i=\Phi(F_i)$ and $ M_j=\Phi(F_j)$, where $M_i, M_j\in\mathbb{R}^{1 \times H \times W}$, 
for indicating the observability at each spatial location.
\begin{figure}[!t]
    \centering
    \includegraphics[width=0.9\columnwidth]{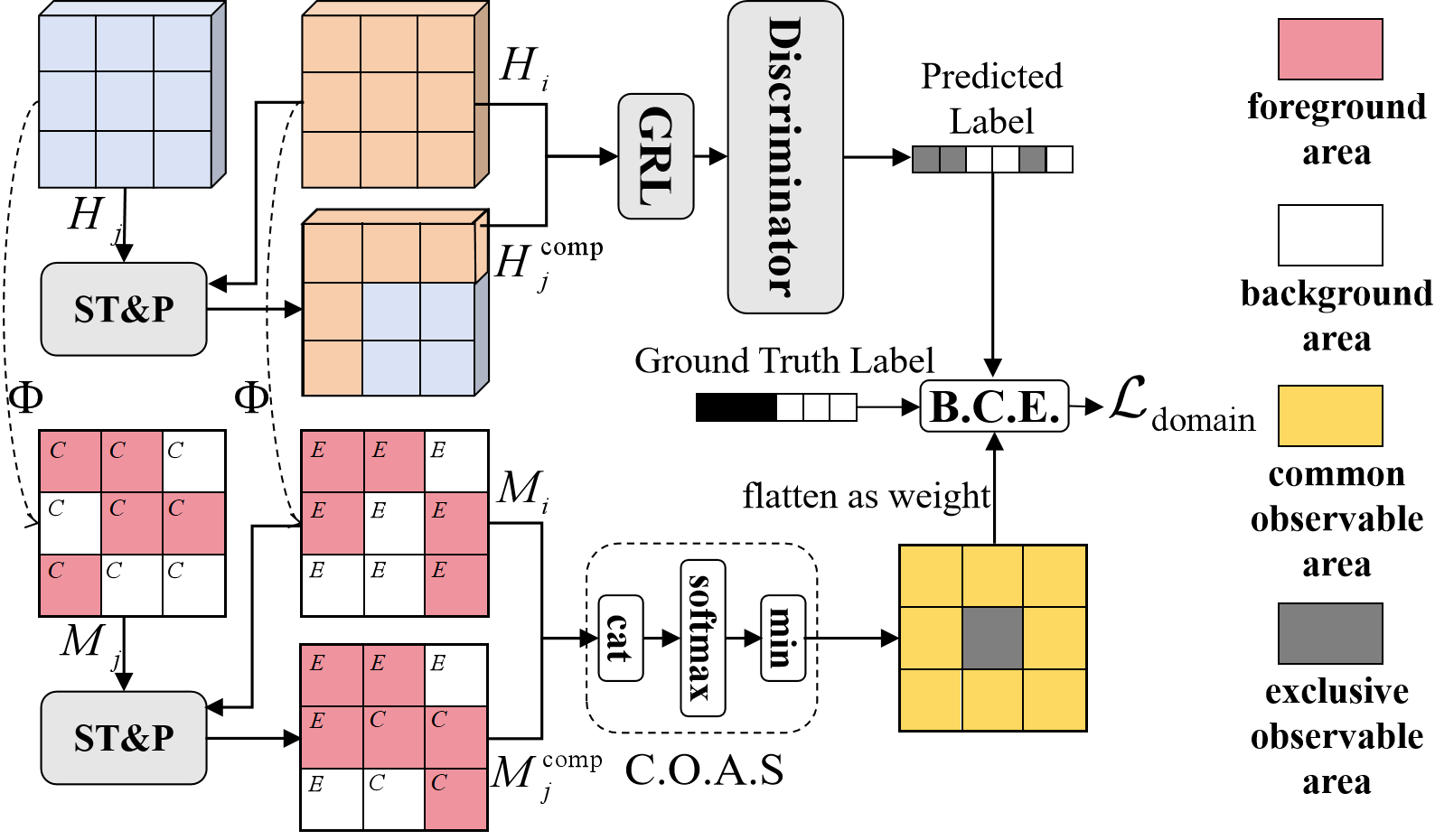}
    \vspace{-0.2cm}
    \caption{Pipeline of OD. ST\&P denotes Spatial Transformation and Padding and C.O.A.S. refers to Common Observable Area Selection. Genuine domain gaps are located by common observable areas, enabling intrinsic domain-invariant feature learning.}
    \label{PipelineofOD}
    \vspace{-0.7cm}
\end{figure}

Subsequently, we project the collaborative features onto the coordinates of ego agent through spatial transformation, creating $H_{j \rightarrow i}$ and $M_{j\rightarrow i}$. However, this may potentially produce void regions. Let ${\cal{V}}$ denote the set of valid grids in the transformed feature map and the voids are filled as
\begin{equation}
    H_j^\text{comp} = {\cal{I}}_{\cal{V}} \cdot H_{j \rightarrow i} + (1-{\cal{I}}_{\cal{V}}) \cdot H_i,
    \label{featcomp}
\end{equation}
\begin{equation}
    M_j^\text{comp} = {\cal{I}}_{\cal{V}} \cdot M_{j \rightarrow i} + (1-{\cal{I}}_{\cal{V}}) \cdot M_i,
    \label{featcomp}
\end{equation}
\noindent where $H_j^\text{comp}$ and $M_j^\text{comp}$ represent the complemented feature and observability map, respectively, and the indicator function ${\cal{I}}_{\cal{V}}$ equals to 1 for points in $V$ and 0 for the others. 
To ensure the domain alignment focus on the regions with shared observability of both agents, an observability weighting map $W\in \mathbb{R}^{1 \times H \times W}$ is computed as $W = \text{min}(\text{softmax}([M_i, M_j^\text{comp}]))$,
where $[\cdot]$ denotes concatenation and all operations are performed along the first dimension. Finally, the domain alignment objective is
{\small{
\begin{equation}
\max_\theta \min_{\mu} \mathcal{L}_{\text{domain}} = \frac{1}{\sum W^{sp}_{\text{flat}}} \sum_{{sp} \in \mathcal{S}} W^{sp}_{\text{flat}} \cdot \mathcal{L}_\text{BCE}(D^{sp}_{\mu}(\Psi_{\theta}), Z^{sp}),
\label{domainloss}
\end{equation}}}
where $\mathcal{S}$ denotes the set of all spatial positions with $sp$ denoting one position, $W_{\text{flat}}$ denotes the flattened observability weighting map,  $\Psi_{\theta}$ is the feature extractor (point clouds as input and BEV features as output), $D_{\mu}$ denotes the discriminator, $\mathcal{L}_\text{BCE}$ is the binary cross-entropy loss, and $Z$ is the domain label (0 for ego agent and 1 for collaborative agent).

To jointly optimize this min-max problem, a gradient reversal layer (GRL) \cite{GRL} is inserted before the discriminator. The GRL leaves the input unchanged in the forward pass and applies a negative scaling factor $\gamma=-0.1$ during backpropagation. This allows the end-to-end adversarial training, where the discriminator learns to distinguish domains in the regions with shared observability and the feature extractor learns to generate domain-invariant features that match physical constraints of multi-agent perception.

\subsection{Progressive Temporal Alignment Module (PTAM)}
\subsubsection{Kinematic Perspective of Temporal Alignment}
Temporal asynchrony causes feature misalignment between agents, presenting a fundamental challenge in multi-agent CP. To compensate the temporal misalignment of collaborative features at the historical time $t-\Delta t$, we leverage a kinematic perspective \cite{flow1,flow2}. The temporal evolution of any-scale features at collaborative agent is formulated as
\begin{equation}
F_j(t,\mathbf{x} + \mathbf{v}(t-\Delta t,\mathbf{x})\Delta t) = F_j(t-\Delta t,\mathbf{x}),
\label{phisics}
\end{equation}
\noindent where $\Delta t$ is a time interval (within the range of typical transmission delay in CP \cite{syncnet, latency1, latency2}), $F_j(t-\Delta t,\mathbf{x})$ denotes the features at position $\mathbf{x}$ of time $t-\Delta t$, and $\mathbf{v}(t-\Delta t,\mathbf{x})$ denotes the velocity field describing the motion of features at time $t-\Delta t$. This captures that features of one position at time $t$ can be obtained by tracing back along the velocity field to the corresponding positions of time $t-\Delta t$.

Following the kinematic formulation, the core mechanism of PTAM realizes feature temporal evolution through three essential components: motion field $\Delta p \in \mathbb{R}^{2 \times H \times W}$ representing $\mathbf{v}$, temporal scaling factor $\xi$ corresponding to $\Delta t$, and displacement compensation operation via bilinear sampling. In the following, we omit $\bf{x}$ for simplicity. 

\subsubsection{Two-Step Implementation for Both Agents}
Based on this core mechanism, the progressive temporal alignment strategy of PTAM consists of a two-stage prediction executed serially. Each stage corresponds to one agent.
{\textbf{(i) First Stage.}} The collaborative agent captures historical motion patterns by utilizing its latest two frames of features $F_j(t-\tau-\Delta T)$ and $F_j(t-\tau)$ to predict an intermediate feature $F_j^\text{inter}$.\\
{\textbf{(ii) Second Stage.}} The ego agent performs adaptive temporal alignment based on received motion information and transmission delay characteristics. Due to the potential communication error between two agents, the received features are defined as $\hat{F}_j(t-\tau)$ and $\hat{F}_j^{\text{inter}}$, which are used to generate temporally aligned features $\hat{F}_j(t)$. To capture hierarchical motion patterns of varying granularities, PTAM processes features simultaneously at three spatial scales. \\
{\textbf{Each Stage Implementation.}}
At each stage with each spatial scale, we execute the core mechanism through two steps, \ie, motion estimation and feature warping.
\begin{figure}[!t]
    \centering
    \includegraphics[width=0.95\columnwidth]{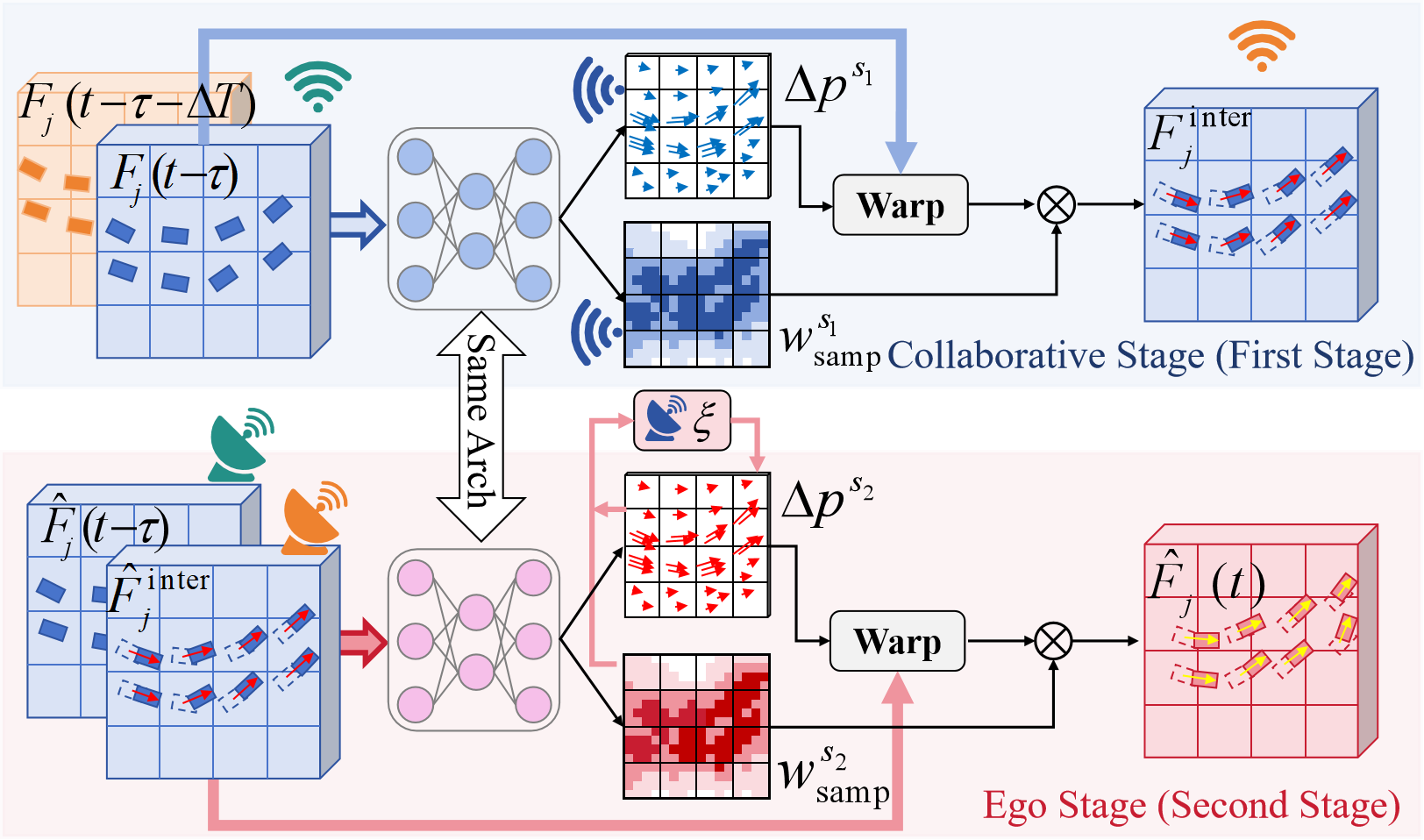}
    \vspace{-0.2cm}
    \caption{Pipeline of PTAM. One spatial scale is illustrated here. Three couples of `WiFi' and `Radar' symbols in different colors denote transmitting and receiving information between agents. Both historical motion tendency and scene dynamics are exploited via two-stage prediction, enabling precise latency compensation.}
    \label{PipelineofPTAM}
    \vspace{-0.8cm}
\end{figure}

\textit{Step 1: Motion Estimation.} This step shares the same architecture for both stages. For a united representation of both stages, we denote the input adjacent temporal features as the latest feature $F_{\text{latest}}$ and its previous feature $F_{\text{prev}}$. The motion estimation begins with computing the temporal feature difference, defined as $\Delta F = F_{\text{latest}} - F_{\text{prev}}$, which is then separately concatenated with these two input features. These concatenated features undergo a sequence of operations to generate the motion field $\Delta p$ and the sampling weight $w_\text{samp} \in \mathbb{R}^{1 \times H \times W}$.

\textit{Step 2: Feature Warping.} This step is slightly different for the two stages.  For the first stage executed at the collaborative agent, a unit temporal scaling factor, \ie, $\xi=1$, is employed to maintain consistency with the sampling period of sensor, e.g., LIDAR, generating an intermediate feature representation $F_j^\text{inter}=w_{\text{samp}}^{s_1} \odot f_\text{warp}\big(F_j(t-\tau), \Delta p^{s_1}\big)$, where  $s_1$ indicates the first stage and $f_\text{warp}(\cdot)$ denotes the bilinear sampling operation. For the second stage at the ego agent, an adaptive temporal scaling factor $\xi$ is required to handle the variable time intervals resulting from transmission delays. The motion difference field $\Delta M$ is defined as
\begin{equation}
\Delta M = (\Delta p^{s_2} \odot w^{s_2}_{\text{samp}}) - (\Delta p^{s_1} \odot w^{s_1}_{\text{samp}}),
\end{equation}
\noindent where the superscript $s_2$ indicates the second stage and the element-wise product captures the effective motion at each stage by incorporating $\Delta p$ and the corresponding $w_{\text{samp}}$. 
Then, to obtain a global context vector $f_M$, we use cascaded convolutional layers, residual blocks, and global pooling operations to process $\Delta M$. And a temporal encoding $f_T$ is obtained by adding sinusoidal positional embeddings, which capture transmission delay characteristics, to $f_M$. Finally, these features are concatenated and processed through a MLP to predict the temporal scaling factor as
\begin{equation}
    \xi = \text{ReLU}(\text{MLP}([f_M, f_T])).
\end{equation}
And the final temporally aligned feature is acquired through $\hat{F}_j(t)=w_{\text{samp}}^{s_2} \odot f_\text{warp}\big(\hat{F}_j^{\text{inter}}, \Delta p^{s_1}\big)$.

\subsubsection{Multi-window Self-supervised Training Strategy.} To enhance the capability of PTAM in capturing diverse motion patterns through progressive-parallel alignment, we propose a multi-window self-supervised training strategy.

At each spatial scale $s$, the feature plane of size $h_s \times w_s$ is partitioned using two complementary window partitioning strategies. These two window sets are defined as
\begin{equation}
    W_1 = \{w_{m',n'} \mid 0 \leq m' < \frac{h_s}{l}, 0 \leq n' < \frac{w_s}{l}\},
    \label{window1}
\end{equation}
\begin{equation}
    W_2 = \{w_{p',q'} \mid 0 \leq p' < \frac{h_s}{l}-1, 0 \leq q' < \frac{w_s}{l}-1\},
    \label{window2}
\end{equation}
\noindent where $w_{m',n'}$ and $w_{p',q'}$ denote windows with top-left corner at position $(m',n')$ and $(p',q')$ with size $l \times l$, respectively. Herein, $W_1$ actually start partitioning features exactly from boundaries while $W_2$ has an offset of $l/2$ compared with $W_1$. 
For each window $w$ in both sets, we compute the cosine similarity $\cos(\cdot,\cdot)$ between predictions and ground truth features $F_{j}^{\text{gt}}$. The loss functions for intermediate and final predictions at scale $s$ can be written as
{\small{
\begin{equation}
{\cal{L}}_\text{inter}^s=\frac{1}{N_\text{window}} \sum_{w \in W_1 \cup W_2} \|1 - \cos(F_{j,w}^{\text{inter},s}, F_{j,w}^{\text{gt},s}(t))\|_2^2,
\label{simloss1}
\end{equation}
\begin{equation}
{\cal{L}}_\text{final}^s=\frac{1}{N_\text{window}} \sum_{w \in W_1 \cup W_2} \|1 - \cos(\hat{F}_{j,w}^s(t), F_{j,w}^{\text{gt},s}(t))\|_2^2,
\label{simloss2}
\end{equation}}}
\noindent where $N_\text{window}$ denotes the total number of windows. Finally, the total temporal alignment loss is computed across all three scales, given by ${\cal{L}}_{\text{temporal}}=\sum_{s=1,2,3}\left({\cal{L}}_{\text{inter }}^s+{\cal{L}}_{\text{final}}^s\right)$.

\subsection{Instance-focused Feature Aggregation Module (IFAM)}

Suppose that  domain-and-time aligned features are obtained by CDAM and PTAM, but how to fully exploit the semantic information of these aligned features is still crucial for improving perception performance. Herein, the IFAM is designed to enhance the structural representation of foreground objects for robust CP. Suppose that $H_a \in \mathbb{R}^{C \times H \times W}$ is the BEV feature of the $a$-th agent, its foreground and background features are identified as $H_a^{\text {fore }}=H_a \odot M_a,\;H_a^{\text {back }}=H_a \odot (\mathbf{1}-M_a)$,
%
where $M_a=\Phi(H_a)$ is the foreground mask and $\Phi(\cdot)$ is the foreground estimator mentioned in Section \ref{OD} (OD module).
\begin{figure}[!t]
    \centering
    \includegraphics[width=\columnwidth]{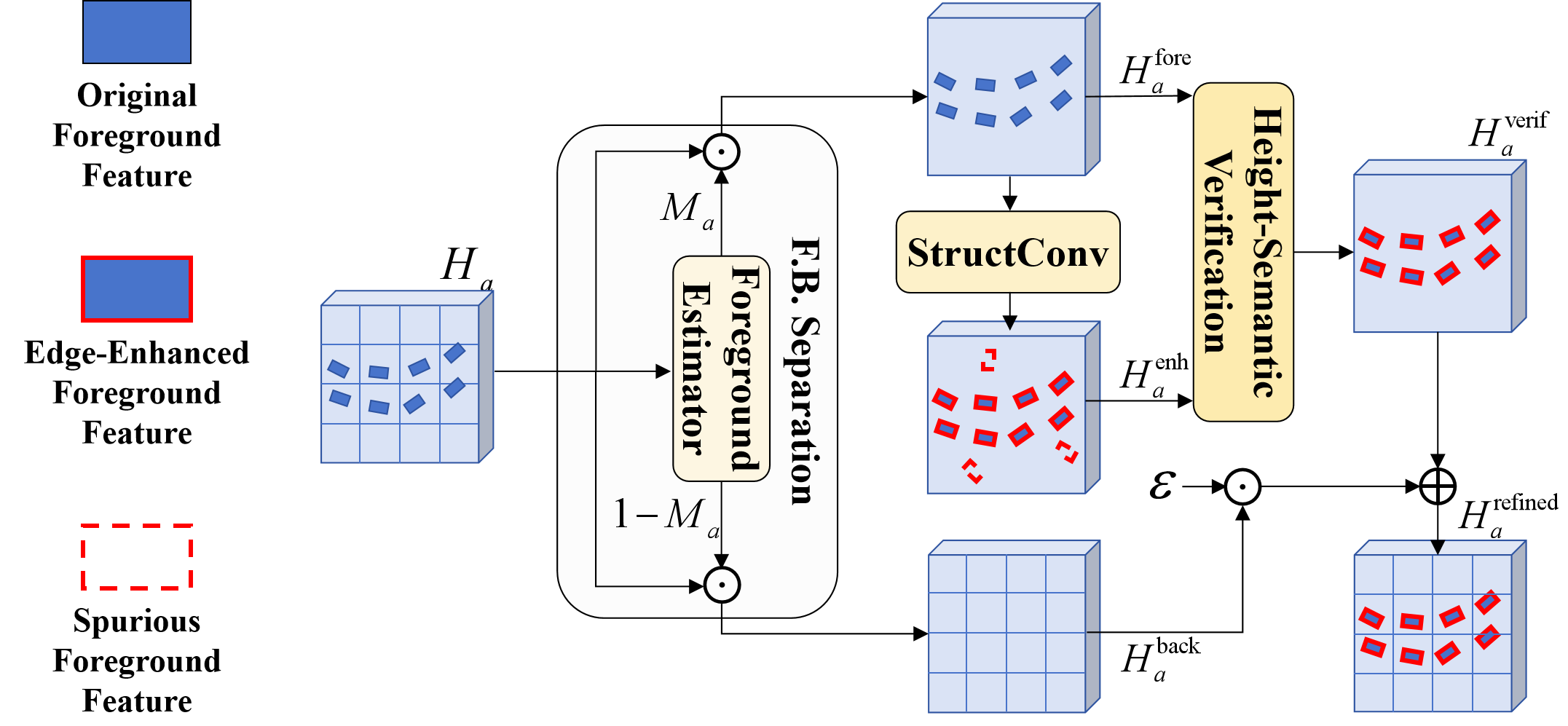}
    \vspace{-0.8cm}
    \caption{Pipeline of IFAM. F.B. denotes fore/background. Structural enhancement benefits from F.B. separation, reducing noise and improving detection via height-semantic verification.}
    \label{PipelineofIFAM}
    \vspace{-0.7cm}
\end{figure}
To strengthen the structural details in foreground regions, a set of $3\times3$ convolution kernels is applied. Then we have $H_a^{\text {enh }}=\text{StructConv}(H_a^{\text {fore }})$, where $\text{StructConv}$ consists of one vanilla convolution to preserve basic feature intensity and four specialized convolutions: a central difference convolution, horizontal and vertical difference convolutions, and an angular difference convolution. Although this strengthens the structural details, it may potentially occur false targets that interfere with subsequent detection.

To suppress these spurious foreground features while preserving enhanced structural details, a foreground verification mechanism is proposed based on the pillar-encoding \cite{pointpillar, pillarnet} since its each channel dimension inherently contains coupled height-semantic information, which can be used to select the correct foreground features. Specifically, given the original and enhanced foreground features ($H_a^\text{fore}$ and $H_a^\text{enh}$), their concatenation along the channel dimension is $H_a^{\text{cat}}=\left[H_a^{\text {fore}}, H_a^{\text{enh}}\right] \in \mathbb{R}^{2 C \times H \times W}$. Then spatial and channel attention modules are employed in parallel. For spatial attention $ W_a^\text{s}$, max- and average-pooling operations are performed across the channel dimension followed by concatenation and convolution. For channel attention $W_a^\text{c}$, spatial average pooling is applied followed by two convolutions.
Then the initial attention weights $W_a^\text{init}$ are derived as $W_a^\text{init}=W_a^\text{s}\oplus W_a^\text{c}$. And the verification weights are
\begin{equation}
    W_a^\text{verif}={\text{Sigmoid}}\left(\operatorname{GConv}\left(\operatorname{CS}\left(\left[H_a^\text{cat}, W_a^\text{init}\right]\right)\right)\right),
    \label{verifiweight}
\end{equation}
\noindent where CS denotes channel shuffle \cite{shufflenet} that promotes cross-group information exchange, breaking fixed channel combinations for comprehensive feature verification. And group convolution  $\text{GConv}(\cdot)$ \cite{resnet} enables independent weight generation for different feature groups, allowing specialized verification of height-semantic patterns.  Through this verification mechanism, the regions exhibiting  height-semantic relationships (\eg, those consistent with typical vehicle shapes and heights) are preserved, while the regions containing unnatural representation combinations of spurious foreground features are mitigated. Then the verified foreground feature of the $k$-th agent ultimately is expressed as 
\begin{equation}
    \begin{aligned}
        H_a^\text{verif}=\operatorname{Conv}_{1\times1}(&(W_a^\text{verif} \odot H_a^\text{fore}\oplus(\mathbf{1}-W_a^\text{verif})\odot H_a^\text{enh}) \\
        &\oplus H_a^\text{fore} \oplus H_a^\text{enh}).
    \end{aligned}
    \label{foreaggr}
\end{equation}
\noindent The individually refined BEV features are obtained by combining background features as $H_a^\text{refined}=H_a^\text{verif} \oplus \epsilon H_a^\text{back}$,
where $\epsilon$ is a learnable parameter balancing the contribution of background features. And the refined features of all $N$ agents are progressively fused through a shared $1 \times 1$ convolution to produce the ultimate CP representation.
\section{Experiments}
\label{sec:Experiments}

\subsection{Datasets and Evaluation Metrics}
Experiments are conducted on three datasets: DAIR-V2X-C \cite{dairv2x}, V2XSET \cite{v2xvit}, and V2XSIM \cite{v2xsim}. Detailed dataset information are presented in Section 2.1 of Supplementary Material. For evaluation, Average Precision (AP) is measured at Intersection-over-Union (IoU) thresholds of 0.50 and 0.70 for the car category on all datasets.

\subsection{Implementation}

The backbone employs PointPillar \cite{pointpillar} architecture with a grid size of 0.4m$\times$0.4m. In CDAM, distance threshold $d_{\text{th}}$ = 50~m and $N_{\text{max}}$ = 2 vehicles are downsampled. The two downsampling ratios are $\beta_{\text{in}}$ = 0.6 and $\beta_{\text{out}}$ = 0.8. For PTAM, the multi-window self-supervised training adopts a window size of $l$ = 16. The training procedure comprises three sequential stages: detection model training, PTAM training, and transmission module training, which are presented in Section 2 of Supplementary Material.

\subsection{Quantitative Results}

\noindent\textbf{Performance without Latency.} Table~\ref{tab:nolatency} presents performance comparison with SOTA methods under zero latency setting. DATA
\begin{table}[!t]
\footnotesize
\setlength{\tabcolsep}{3pt}
\begin{tabular}{@{}lccc@{}}
\toprule
\multirow{1}{*}{Method $\|$ Metric(AP$_{50}$/AP$_{70}$)} & V2XSim & V2XSet & DAIR-V2X-C \\
\midrule
DiscoNet (NeurIPS'21) \cite{disconet}  & 83.56 / 66.12 & 82.34 / 64.79 & 68.50 / 53.57\\
AttFuse (ICRA'22) \cite{opv2v} & 81.70 / 66.24 & 84.37 / 66.27 & 67.36 / 52.96 \\
V2X-VIT (ECCV'22) \cite{v2xvit} & 82.32 / 64.41 & 82.42 / 63.14 & 71.54 / 51.65 \\
CoBEVT (CoRL'22)  \cite{cobevt}& 81.00 / 65.06 & 84.84 / 65.14 & 69.21 / 46.66 \\
Where2com (NeurIPS'22) \cite{where2comm} & 83.82 / 65.52 & 85.19 / 61.59 & 68.39 / 52.48 \\
AdaFusion (WACV'23)  \cite{adafusion}& 78.89 / 58.62 & 86.28 / 57.06 & 71.16 / 47.74 \\
HM-VIT (ICCV'23)  \cite{hmvit} & - / - & - / - & 76.10 / - \\
CoBEVFlow (NeurIPS'23)  \cite{cobevflow}& - / - & - / - & 73.80 / 59.90 \\
FFNet (NeurIPS'23)  \cite{ffnet}& 85.56 / 68.64 & 83.57 / 66.23 & 77.19 / 60.17 \\
DI-V2X (AAAI'24) \cite{div2x} & - / - & - / - & 78.82 / - \\
MRCNet (CVPR'24) \cite{mrcnet} & 85.33 / 69.82 & 85.00 / 66.31 & - / - \\
CodeFilling (CVPR'24) \cite{codefilling} & - / - & - / - & 79.90 / 61.06 \\
HEAL (ICLR'24)  \cite{heal}& 88.67 / 75.85 & 88.40 / 72.13 & 79.00 / 63.12 \\
\midrule
DATA (Ours) & \textbf{88.91} / \textbf{77.69} & \textbf{89.53} / \textbf{74.98} & \textbf{79.94} / \textbf{65.48} \\
\bottomrule
\end{tabular}
\vspace{-0.3cm}
\caption{Performance comparison without communication latency.}
\label{tab:nolatency}
\vspace{-0.6cm}
\end{table}
achieves 0.94\% and 2.36\% improvements in AP$_{50}$ and AP$_{70}$ compared to HEAL \cite{heal} on real-world DAIR-V2X-C validation set. For simulation datasets V2XSIM and V2XSET, where Gaussian noise ($\mu_\text{pos}=0, \sigma_\text{pos}=0.2~m$ for position, $\mu_\text{rot}=0,\sigma_\text{rot}=0.2^{\circ}$ for orientation) is added to mimic real-world conditions, DATA also demonstrates significant improvements. Specifically, it outperforms MRCNet, by 3.58\% and 7.87\% in AP$_{50}$ and AP$_{70}$ on V2XSIM, and by 4.53\% and 8.67\% on V2XSET. These consistent improvements across both complex simulated and real-world scenarios demonstrate the effectiveness of DATA, which first achieves domain-invariant feature extraction through CDAM to enhance feature quality, then leverages IFAM to effectively highlight foreground feature semantics for improved CP system performance.

\begin{figure}[!t]
    \centering
    \includegraphics[width=0.9\columnwidth]{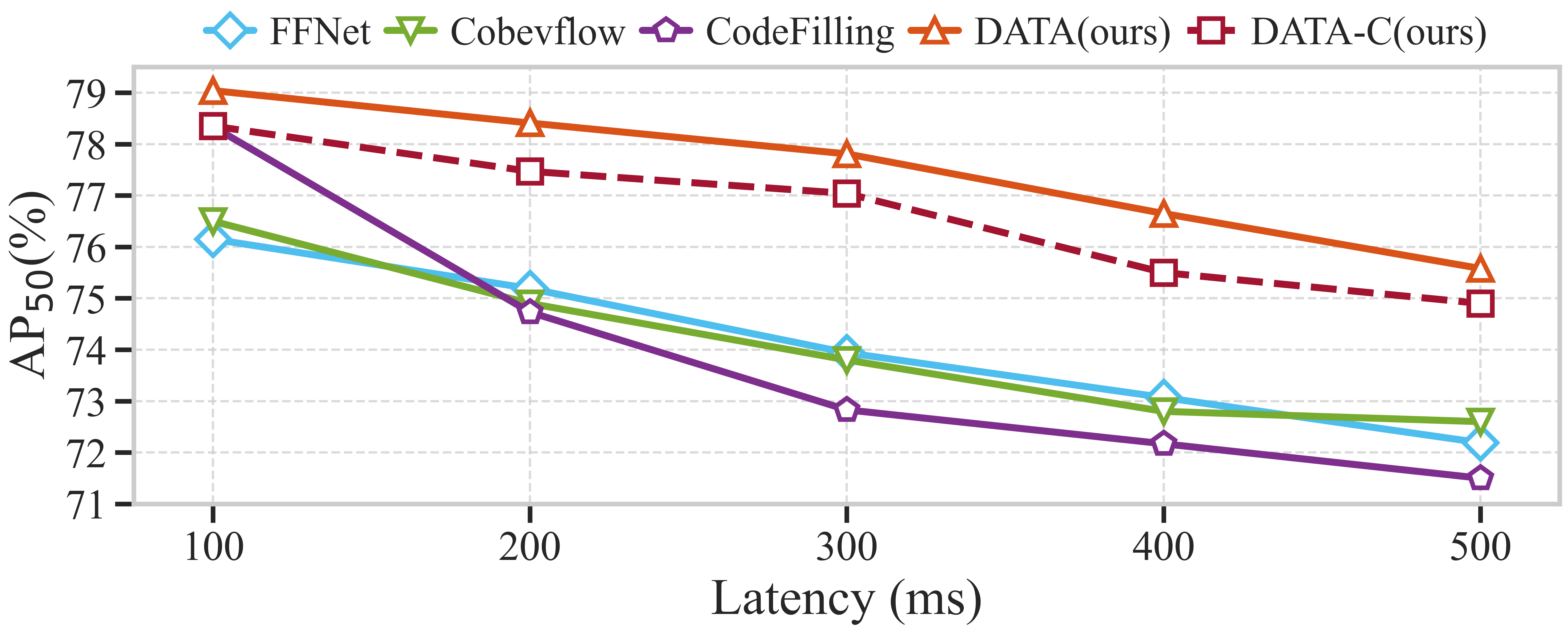}
    \vspace{-0.4cm}
    \caption{Performance with latency on DAIR-V2X-C.}
    \label{fig:robustlatency}
    \vspace{-0.8cm}
\end{figure}

\noindent\textbf{Performance with Latency.} As shown in Figure~\ref{fig:robustlatency}, DATA effectively maintains high-quality CP under temporal asynchrony on DAIR-V2X-C dataset. Building upon the strong feature extraction and fusion capabilities in zero-latency scenarios, the PTAM further ensures reliable feature prediction during transmission with varying delays, realizing the minimal 3.46\% AP$_{50}$ decrease as latency increases from 100ms to 500ms. And DATA maintains 75.58\% AP$_{50}$ at 500ms delay, while baseline methods, like FFNet (72.19\%) and CodeFilling (71.50\%), present more significant drops. Also, DATA achieves robust performance even with compressed transmission (74.90\% AP$_{50}$ at 500ms of DATA-C).
These results demonstrate that DATA effectively addresses temporal challenges in real-world CP by acquiring and utilizing high-quality features.

\begin{figure}[!t]
    \centering
    \includegraphics[width=0.9\columnwidth]{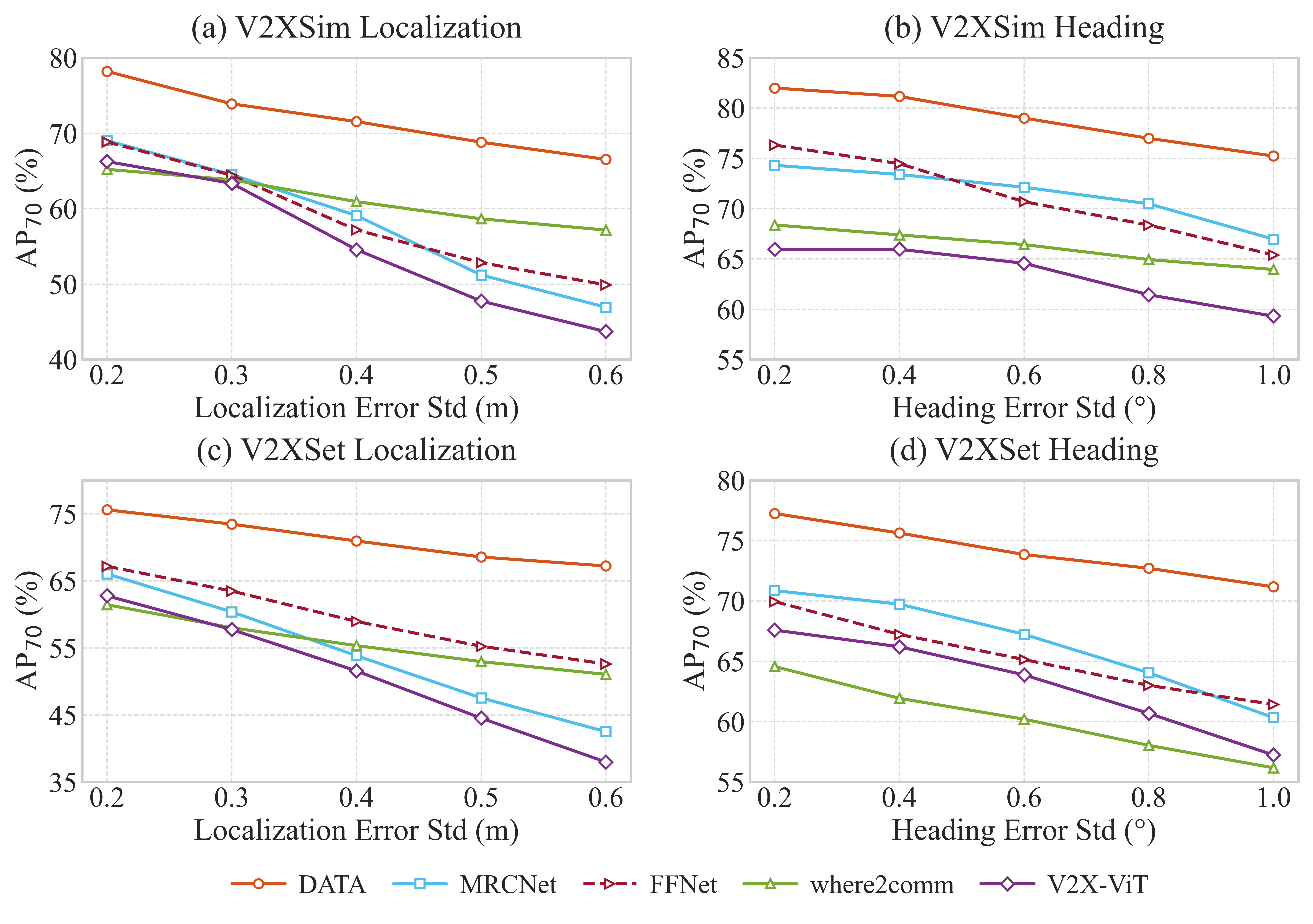}
    \vspace{-0.4cm}
    \caption{Performance comparison with pose errors.}
    \label{fig:robustnoise}
    \vspace{-0.6cm}
\end{figure}

\noindent\textbf{Robustness to Pose Errors.} To investigate the effects of different pose noise components on CP performance, experiments are conducted on V2XSET and V2XSIM. Gaussian noise with varying standard deviations is separately applied to collaborator positions and orientations, with localization noise $\sigma_\text{loacl}$ ranging from $0.2~m$ to $0.6~m$ and heading noise $\sigma_\text{head}$ ranging from $0.2^{\circ}$ to $1.0^{\circ}$. As shown in Figure~\ref{fig:robustnoise}, the compared methods suffer from significant performance degradation. Their AP$_{70}$ declines rapidly with increasing noise magnitude, dropping by over 30\% and 9\% when $\sigma_\text{local}$ and $\sigma_\text{head}$ reaches $0.6~m$ and $1.0^{\circ}$, respectively, on both V2XSIM and V2XSET. This indicates that spatial feature misalignment severely impacts their CP accuracy. DATA demonstrates robust resilience against pose uncertainties on both datasets: on V2XSIM, it maintains 66.53\% AP$_{70}$ at $\sigma_{\text{local}}=0.6~\text{m}$ and 75.23\% AP$_{70}$ at $\sigma_{\text{head}}=1.0^{\circ}$; similarly on V2XSET, it achieves 67.22\% AP$_{70}$ at $\sigma_{\text{local}}=0.6~\text{m}$ and 71.17\% AP$_{70}$ at $\sigma_{\text{head}}=1.0^{\circ}$. This robustness stems from two complementary aspects. First, the observability-guided domain alignment learns domain-invariant features from consistent observation, extracting essential geometric patterns from noisy observations. Meanwhile, the instance-focused feature enhancement, through foreground-background separation and semantic refinement, strengthens the structural completeness of object representations, ensuring reliable detection under pose perturbations.

\subsection{Ablation Study}

\noindent\textbf{Ablation on Domain Alignment and Feature Aggregation.} To investigate the contribution of each module, experiments are conducted on DAIR-V2X-C validation set without latency. As shown in Table~\ref{tab:ablation_domain}, compared with the baseline model, incorporating IFAM brings significant performance gains, \ie, 3.70\% improvement in AP$_{70}$, since its geometric-aware feature enhancement strengthens structural completeness and enables more precise object localization. Adding OD further improves the detection performance by 1.96\% AP$_{50}$ and 0.62\% AP$_{70}$ since it guides the feature alignment to focus on regions where both agents maintain comparable observation capabilities, fundamentally addressing cross-domain feature discrepancies. The integration of PHD yields 0.96\% and 1.25\% improvements in AP$_{50}$ and AP$_{70}$. This demonstrates that maintaining consistent distribution of point clouds across different spatial regions, while preserving inherent scene structures, is crucial for extracting reliable domain-invariant features.

\noindent\textbf{Ablation on Proximal Region Threshold.} The PHD module aims to balance point clouds density across different spatial regions while preserving scene structure. The results in Table \ref{tab:distance_threshold_comparison} demonstrate that the $50~m$ threshold achieves optimal performance (79.94\% AP$_{50}$), representing a good trade-off between density balancing and distribution preservation. Smaller thresholds ($10~m, 30~m$) provide insufficient coverage of proximal objects, while larger thresholds ($70~m, 90~m$) may introduce excessive downsampling that disrupts the original scene structures. 

\noindent\textbf{Ablation on Progressive Temporal Alignment.} To investigate the impact of window size on temporal alignment performance, experiments are conducted, comparing  First-Stage and Two-Stage models, on DAIR-V2X-C validation set with 300~ms communication delay. Table \ref{tab:window_length_comparison} shows that the Two-stage model outperforms the First-stage model for all window sizes, where the Two-Stage model achieves optimal performance (77.81\% AP$_{50}$) with window size 16, surpassing the whole map supervision and the First-Stage approach by 0.67\% and 1.01\%, respectively. 
The performance degrades when increasing or decreasing the window size, indicating that this intermediate window size optimally balances local motion capture and contextual information preservation. Especially with small window sizes, like 4, both models face challenges since small windows 
cannot adequately capture the vehicle objects. And the Two-Stage model shows more significant performance degradation (76.64\% AP$_{50}$ at size 4) compared to the One-Stage approach (76.54\% AP$_{50}$) due to the additional impact of error accumulation through prediction stages. These findings validate that Two-Stage progressive alignment can achieve temporal coherence through intermediate prediction states when trained with appropriate window partitions.
\begin{table}[!t]
\centering
\small
\renewcommand{\tabcolsep}{1pt}
\begin{tabular}{@{}l@{\hspace{2pt}}c@{\hspace{2pt}}c@{\hspace{4pt}}l@{\hspace{2pt}}c@{\hspace{2pt}}c@{}}
\toprule
\multirow{1}{*}{Config} & AP$_{50}$ & AP$_{70}$ & \multirow{1}{*}{Config} & AP$_{50}$ & AP$_{70}$ \\
\midrule
Baseline & 76.80 & 59.91 & + P. + O. & 78.96$^{\tiny\textcolor{red}{+2.16}}$ & 62.68$^{\tiny\textcolor{red}{+2.77}}$ \\
+ P. & 77.54$^{\tiny\textcolor{red}{+0.74}}$ & 62.16$^{\tiny\textcolor{red}{+2.25}}$ & + P. + I. & 78.30$^{\tiny\textcolor{red}{+1.50}}$ & 63.65$^{\tiny\textcolor{red}{+3.74}}$ \\
+ O. & 78.52$^{\tiny\textcolor{red}{+1.72}}$ & 62.58$^{\tiny\textcolor{red}{+2.67}}$ & + O. + I. & 78.98$^{\tiny\textcolor{red}{+2.18}}$ & 64.23$^{\tiny\textcolor{red}{+4.32}}$ \\
+ I. & 77.02$^{\tiny\textcolor{red}{+0.22}}$ & 63.61$^{\tiny\textcolor{red}{+3.70}}$ & + P. + O. + I. & \textbf{79.94}$^{\tiny\textcolor{red}{+3.14}}$ & \textbf{65.48}$^{\tiny\textcolor{red}{+5.57}}$ \\
\bottomrule
\end{tabular}
\vspace{-0.3cm}
\caption{Comprehensive ablation study of CDAM and IFAM on DAIR-V2X-C. P.: PHD, O.: OD, I.: IFAM.}
\label{tab:ablation_domain}
\vspace{-0.4cm}
\end{table}
\begin{table}[!t]
\centering
\small
\renewcommand{\tabcolsep}{1.7pt}
\begin{tabular}{@{}lcccccc@{}}
\toprule
\multirow{2}{*}{Model Type (AP$_{50}$ \%)} & \multirow{2}{*}{\begin{tabular}[c]{@{}c@{}}w/o\\PTAM\end{tabular}} & \multicolumn{5}{c}{Window Size $l$ (AP$_{50}$ \%)} \\
\cmidrule(lr){3-7}
& & w.m. & 32 & 16 & 8 & 4 \\
\midrule
First-Stage (Collab.) & 71.65 & 76.23 & 76.68 & 76.80 & 76.62 &  76.54 \\
Two-Stage (Ego \& Collab.) & 71.65 & 77.14 & 77.62 & \textbf{77.81} & 77.59 & 76.64 \\
\bottomrule
\end{tabular}
\vspace{-0.3cm}
\caption{Performance on 300~ms latency of Two model types across different window sizes. ``w.m.'' represents whole map supervision.}
\label{tab:window_length_comparison}
\vspace{-0.5cm}
\end{table}

\begin{table}[!t]
\centering
\small
\renewcommand{\tabcolsep}{6pt}
\begin{tabular}{@{}l@{\hspace{4pt}}ccccc@{}}
\toprule
Distance Threshold $d_{th}$ & 10m & 30m & 50m & 70m & 90m \\
\midrule
Metric (AP$_{50}$ \%) & 79.33 & 79.44 & \textbf{79.94} & 79.18 & 78.69 \\
\bottomrule
\end{tabular}
\vspace{-0.3cm}
\caption{Ablation of proximal-region distance thresholds ($d_{th}$).}
\label{tab:distance_threshold_comparison}
\vspace{-0.8cm}
\end{table}

\section{Conclusion}
\label{sec:Conclusion}
In this paper, we propose DATA, which is a novel framework that addresses the fundamental domain and time misalignment in feature-level collaborative perception. 
We develop systematic solutions for feature alignment during the acquisition phase. The proposed CDAM reduces domain gaps in feature extraction through density-aware point cloud sampling and observability-guided domain alignment, while PTAM ensures temporal coherence during feature transmission through progressive motion refinement. Building upon well-aligned features, IFAM further enhances their semantic expressiveness to maximize fusion performance. Extensive experiments on both real-world and simulation datasets demonstrate the effectiveness of DATA, achieving SOTA performance under various challenging conditions, validating the robustness of our systematic approach on feature alignment and enhancement in CP.
\section*{Acknowledgments}
This work was supported in part by the National Natural Science Foundation of China, in part by the Nanjing U35 Strong Foundation Engineering, in part by the Key Research and Development Program of Jiangsu Province.

{
    \small
    \bibliographystyle{ieeenat_fullname}
    \bibliography{main}
}

\clearpage
\setcounter{page}{1}
\setcounter{section}{0}
\maketitlesupplementary

\section{Detailed Information about Module Designs}

\subsection{Observability-constrained Discriminator Implementation}

The observability-constrained discriminator (OD) employs a lightweight convolutional architecture to distinguish between ego and collaborative domains. The discriminator consists of two convolutional layers: a first layer with 256 channels and kernel size $1 \times 1$ followed by ReLU activation, and a final layer with a single output channel and kernel size $1 \times 1$ that outputs predicted domain labels.


\subsection{Multi-scale Feature Processing for BEV Representation}

The multi-scale feature processing module transforms features at different spatial resolutions to form a comprehensive Bird's Eye View (BEV) representation.

Assume that three scales of features, which are denoted as $F_{\text{large}} \in \mathbb{R}^{64 \times H \times W}$, $F_{\text{middle}} \in \mathbb{R}^{128 \times \frac{H}{2} \times \frac{W}{2}}$, and $F_{\text{small}} \in \mathbb{R}^{256 \times \frac{H}{4} \times \frac{W}{4}}$, are extracted from the backbone network, and then the conversion to BEV features follows a systematic process.

Each scale undergoes a deconvolution operation to unify spatial and channel dimensions, and this operation can be expressed as
\begin{equation}
    U_{\text{large}} = \Theta_1(F_{\text{large}}),
    \label{bev1}
\end{equation}
\begin{equation}
{U}_{\text{middle}} = \Theta_2({F}_{\text{middle}}),
    \label{bev2}
\end{equation}
\begin{equation}
{U}_{\text{small}} = \Theta_3({F}_{\text{small}}),
    \label{bev3}
\end{equation}
\noindent where $\Theta_1$, $\Theta_2$, and $\Theta_3$ represent deconvolution networks with strides 1, 2, and 4, respectively. Each deconvolution network consists of a transposed convolution layer, batch normalization, and ReLU activation, projecting the features to a uniform spatial resolution of $H \times W$ and channel dimension of 128.

The final BEV representation is formed by concatenating these upsampled features along the channel dimension, formulated as $H_\text{BEV} = [U_{\text{large}}, U_{\text{middle}}, U_{\text{small}}]$. The resulting BEV feature map ${H_\text{BEV}} \in \mathbb{R}^{384 \times H \times W}$ integrates information from all scales while maintaining consistent spatial dimensions. This approach, which is essential for accurate 3D object detection in the BEV space, preserves both fine-grained details from larger-scale features and semantic context from smaller-scale features.

\subsection{Computational Complexity Analysis: Block-Wise vs. Global Computation}

We analyze the computational complexity of block-wise computation versus global computation for calculating cosine similarity between feature maps. Consider feature maps of size $H \times W$ with $C$ channels, then for the block-wise approach, we use the window size $l \times l$, where $l$ is divisible by both $H$ and $W$ for simplicity. Herein, $\mathbf{f}_{\text{p}}$ denotes the predicted feature map and $\mathbf{f}_{\text{gt}}$ denotes the ground truth feature map.

\subsubsection{Global Computation}
For global computation across the entire feature map, the computational cost is listed as follows:
\begin{description}[
    labelindent=0em,
    labelwidth=1.5em,
    labelsep=0.5em,
    leftmargin=2em,
    align=left,
    font=\normalfont
]
\item[(i)] Computing vector dot products $\langle \mathbf{f}_{\text{p}}, \mathbf{f}_{\text{gt}} \rangle$ requires $C \times H \times W$ multiplications and $(C - 1) \times H \times W$ additions.
\item[(ii)] Computing $\ell_2$ norms $\|\mathbf{f}_{\text{p}}\|_2$ and $\|\mathbf{f}_{\text{gt}}\|_2$ requires $2 \times C \times H \times W$ multiplications for squaring each element, $2 \times (C-1) \times H \times W$ additions for summing squared elements, and $2 \times H \times W$ square root operations.
\item[(iii)] Computing cosine similarity $s =\frac{\langle \mathbf{f}_{\text{p}}, \mathbf{f}_{\text{gt}} \rangle}{\|\mathbf{f}_{\text{p}}\|_2 \cdot \|\mathbf{f}_{\text{gt}}\|_2}$ requires $H \times W$ divisions and $H \times W$ multiplications.
\item[(iv)] Finally, computing MSE loss $(s - 1)^2$ requires $H \times W$ subtractions, $H \times W$ squaring operations, and $(H \times W - 1)$ additions for the final sum.
\end{description}



In total, the global computation requires approximately $(3C + 2) \times H \times W$ multiplications, $(3C - 1) \times H \times W$ additions, $2 \times H \times W$ square roots, and $H \times W$ divisions, yielding a computational complexity of $\mathcal{O}(C \times H \times W)$.

\subsubsection{Block-Wise Computation with Multi-Window Strategy}
Our approach uses two complementary window partitioning strategies:

The first strategy ($W_1$ described in Eq. (11)) applies standard partitioning, yielding $\lfloor H/l \rfloor \times \lfloor W/l \rfloor$ windows. The second strategy ($W_2$ described in Eq. (12)) employs offset partitioning (by $l/2$ compared with $W_1$), yielding $\lfloor (H-l)/l \rfloor \times \lfloor (W-l)/l \rfloor$ windows due to the offset introduced along the top, bottom, left, and right boundaries of the feature map. For typical dimensions in our implementation ($H = 256, W = 128, l = 16$), we have $|W_1|= 16 \times 8 = 128$ windows and $|W_2|= 15 \times 7 = 105$ windows.

For each window in either strategy, computing dot products requires $C \times l \times l$ multiplications and $(C-1) \times l \times l$ additions. $\ell_2$ norm computation requires $2 \times C \times l \times l$ multiplications, $2 \times (C-1) \times l \times l$ additions, and $2 \times l \times l$ square roots. Computing cosine similarity requires $l \times l$ divisions and $l \times l$ multiplications, and MSE loss computation requires $l \times l$ subtractions and $l \times l$ multiplications.

The total computation across both window strategies is 
\begin{equation}
\begin{aligned}
&\left[\lfloor\frac{H}{l}\rfloor \times \lfloor\frac{W}{l}\rfloor + \right.\left. \lfloor\frac{H-l}{l}\rfloor \times \lfloor\frac{W-l}{l}\rfloor\right] \times \\
&[(3C+2) \times l \times l].
\end{aligned}
\label{Oofwindow}
\end{equation}
\noindent With our typical dimensions, this is approximately $1.8 \times (3C+2) \times H \times W$, giving a computational complexity of $\mathcal{O}(C \times H \times W)$. This is asymptotically equivalent to the global approach.

\subsubsection{Advantages of Block-Wise Computation}

The block-wise approach with dual window strategy offers several significant advantages in temporal feature alignment as follows:

First, the windowed computation provides stronger learning signals for capturing localized motion patterns, enabling the model to better distinguish between different motion behaviors within the same scene. Traffic scenes exhibit both structured patterns and individual object movements, which benefit from this localized analysis approach.

Second, window-based computation rebalances the influence of foreground objects in similarity calculations, effectively counteracting the predominant impact of background regions that typically dominate the feature space. Consider that an object spans $m \times n$ pixels, where $m, n \textless l$. In global similarity computation, the computation of foreground objects would be proportional to $(m \times n)/(H \times W)$ when calculating the overall cosine similarity metric. However, in the window-based approach, where similarity is calculated independently for each window before computing the MSE loss, the foreground objects make more contribution to the similarity, reaching $(m \times n)/(l \times l)$. This would become even more significant when window size $l$ is smaller than the feature dimensions $H$ and $W$. This enhanced representation of foreground elements strengthens motion pattern modeling for salient objects, which typically occupy smaller spatial regions compared to the background.


Third, the complementary window partitioning strategies ensure comprehensive spatial coverage. Given the window size $l$ and the offset $l/2$, any object of size up to $l \times l$ will be fully contained within at least one window from either partitioning strategy, regardless of its position in the feature map. This property prevents foreground objects from being fragmented across multiple windows, providing coherent supervision signals for learning object motion patterns.

Fourth, the block-wise approach addresses the challenge of balancing attention between foreground and background regions. By processing feature maps in window-based units, the model allocates more balanced computational attention across the feature space, rather than being dominated by background regions that typically occupy the majority of the scene. As demonstrated in our experimental results, this approach significantly improves temporal alignment quality and detection performance, particularly for foreground objects of interest in autonomous driving scenarios.

\subsection{StructConv: Multi-directional Structure Enhancement}

The StructConv operation employs five specialized $3 \times 3$ convolutions that collectively enhance structural details in foreground features. Each targets specific geometric patterns while maintaining computational efficiency through parameter combination. We detail these specialized convolutions as follows.

\textbf{Feature Preservation Convolution.} This convolution maintains the original feature representation by computing standard weighted sums across neighborhood pixels. This serves as the foundation upon which structural enhancements are built while preserving essential intensity information.

\textbf{Center-Surround Contrast Convolution.} This convolution enhances local contrast by modifying the center weight of the kernel to be the negative sum of all surrounding weights. This operation highlights intensity transitions at feature boundaries and enhances object contours.

\textbf{Horizontal Edge Convolution.} This convolution detects horizontal structures by placing positive weights on the left column, zeros in the center column, and exact negatives of the left-side weights on the right column of the kernel. This symmetrical weight pattern creates a directional gradient detector that emphasizes horizontal edges commonly found in vehicle appearance.

\textbf{Vertical Edge Convolution.} This convolution captures vertical features through a principle similar to the Horizontal Edge Convolution, but with positive weights at the top row, zeros in the middle row, and exact negatives of the top-row weights at the bottom row of the kernel. This arrangement enhances vertical boundaries and surface transitions in the scene.

\textbf{Diagonal Structure Convolution.} This convolution identifies corner features and angular transitions through a specialized weight permutation. The implementation subtracts a rotated version of the original kernel from itself, creating a pattern that responds strongly to diagonal intensity gradients present at object corners.

While these convolutions effectively enhance structural details, they may introduce spurious foreground features that may negatively impact detection. To address this, the enhanced features undergo a coupled height-semantic verification process that selectively preserves structural enhancements while suppressing false features. This verification leverages the inherent channel-wise height and semantic correlations in pillar-encoded features to ensure that only structurally reasonable enhancements are retained.

The five convolutions are efficiently implemented by combining their weights and biases into a single operation, providing comprehensive structural enhancement without the computational overhead of separate convolutions.

\subsection{Foreground Estimator Implementation}

The foreground estimator $\Phi(\cdot)$, utilized in both CDAM and IFAM, is implemented as a lightweight network to identify foreground regions. The network architecture consists of a $3 \times 3$ convolution that halves the input channel dimension, followed by batch normalization and ReLU activation, and a final $1 \times 1$ convolution with sigmoid activation to produce single-channel output.

Herein, $\Phi(\cdot)$ is supervised by the foreground occupancy loss function that computes a weighted focal loss between predicted foreground maps and foreground labels across spatial locations. This loss function emphasizes the importance of correctly identifying occupied regions while accounting for the class imbalance between foreground and background pixels.

Then the foreground occupancy loss ${\cal{L}}_{foreground}$ can be formulated as
\begin{equation}
    \mathcal{L}_{foreground} = \sum_{s=1}^{S} w_{s} \cdot \mathcal{L}_{focal}(M_{s}, M_{s}^\text{gt}),
    \label{foreloss}
\end{equation}
\noindent where $S$ is the total number of spatial locations in the feature map, $M_{s}$ denotes the predicted foreground value from the foreground estimator at location $s$, and $M_{s}^\text{gt}$ is the corresponding ground truth foreground label derived from bounding boxes.

The weighting factor $w_{s}$ is defined as
\begin{equation}
    w_{s} = \frac{M_{s}^\text{gt} \cdot w_{pos} + (1-M_{s}^\text{gt}) \cdot w_{neg}}{\max\left(\sum_{s=1}^{S} M_{s}^\text{gt}, 1\right)},
\end{equation}
\noindent where $w_{pos}=2$ is the weight for positive samples and $w_{neg}=1$ is the weight for negative samples. The denominator normalizes the weights by the total number of positive samples, with a minimum value of 1 to prevent division by zero.

The focal loss $\mathcal{L}_{focal}$ can be calculated as
\begin{equation}
\begin{aligned}
    \mathcal{L}_{focal}(p, y) = &-(y \cdot 0.25 \cdot (1-p)^2 \cdot \log(p)\\
    &+ (1-y) \cdot (1-\alpha) \cdot p^2 \cdot \log(1-p)).
\end{aligned}
\end{equation}

This supervisory signal guides the foreground estimator to accurately predict foreground occupancy while handling the inherent imbalance between foreground and background regions in the scene.

\section{Detailed Dataset Information and Experiment Configuration }\label{suppexp}

\subsection{Detailed Dataset Information}\label{datasetdetail}
Experiments are conducted on three collaborative perception datasets: DAIR-V2X-C \cite{dairv2x}, V2XSET \cite{v2xvit}, and V2XSIM \cite{v2xsim}. (i) DAIR-V2X-C, which focuses on vehicle-infrastructure collaborative data, is a subset of DAIR-V2X and contains 39,000 frames. Following the official benchmark in \cite{dairv2x}, a synchronized subset VIC-Sync with 9,311 frame pairs is split into training/validation/testing sets in a ratio of 5:2:3. We adopt the complemented annotations \cite{coalign}, with the perception range is set to $x \in [-102.4m, 102.4m]$ and $y \in [-51.2m, 51.2m]$. (ii) V2XSIM is a cooperative perception dataset co-simulated by SUMO \cite{sumo} and CARLA \cite{carla}. The dataset consists of 10,000 frames with 501K annotated boxes, split into 8,000/1,000/1,000 for training/validation/testing. The perception range is set to $x \in [-32m, 32m]$ and $y \in [-32m, 32m]$. (iii) V2XSet, co-simulated by OpenCDA \cite{opencda} and CARLA, is a V2X dataset with realistic noise simulation. The dataset contains 11,447 frames, split into 6,694/1,920/2,833 for training/validation/testing. The perception range is set to $x \in [-140m, 140m]$ and $y \in [-40m, 40m]$.

\subsection{Three-stage Training and Loss Function}\label{3stagetrain}

The DATA framework employs a sequential three-stage training strategy, each with specialized loss functions tailored to specific aspects of the model.

\subsubsection{Stage 1: Domain Alignment and Feature Extraction}
In the first stage, we jointly optimize the encoder, CDAM, IFAM, and detection head by using synchronized data. The total loss function for this stage is formulated as
\begin{equation}
\mathcal{L}_{\text{stage1}} = \mathcal{L}_{\text{det}} + \lambda_{\text{fore}} \cdot \mathcal{L}_{\text{foreground}} + \lambda_{\text{domain}} \cdot \mathcal{L}_{\text{domain}}
\end{equation}
where $\mathcal{L}_{\text{det}}$ is the detection loss comprising three components as
\begin{equation}
\mathcal{L}_{\text{det}} = \mathcal{L}_{\text{cls}} + \lambda_{\text{reg}} \cdot \mathcal{L}_{\text{reg}} + \lambda_{\text{dir}} \cdot \mathcal{L}_{\text{dir}},
\end{equation}
\noindent where $\mathcal{L}_{\text{cls}}$ denotes a Sigmoid Focal Loss for classification, $\mathcal{L}_{\text{reg}}$ represents a Weighted Smooth ${\ell_1}$ Loss for regression, and $\mathcal{L}_{\text{dir}}$ indicates a Weighted Softmax Classification Loss for direction prediction. The weighting factors are set to $\lambda_{\text{reg}}=2.0$, $\lambda_{\text{dir}}=0.2$, $\lambda_{\text{fore}}=0.4$, and $\lambda_{\text{domain}}=1.0$.

\subsubsection{Stage 2: Temporal Alignment}
The second stage focuses on training PTAM with asynchronous data while freezing all parameters from the first stage. The loss function is defined as
\begin{equation}
\mathcal{L}_{\text{stage2}} = \mathcal{L}_{\text{det}} + \lambda_{\text{temporal}} \cdot \mathcal{L}_{\text{temporal}},
\end{equation}
where the detection loss $\mathcal{L}_{\text{det}}$ maintains the same form and parameters as in Stage 1, while $\mathcal{L}_{\text{temporal}}$ is the multi-scale and multi-window temporal alignment loss described at the end of Section 3.3.3. Herein, we set $\lambda_{\text{temporal}}=1.0$ to balance the influence of detection and temporal alignment objectives.

\subsubsection{Stage 3: Compression Network}
The final stage trains the transmission compression and recovery network by using asynchronous data with all parameters from previous stages frozen. The loss function combines detection objectives with reconstruction fidelity as
\begin{equation}
\mathcal{L}_{\text{stage3}} = \mathcal{L}_{\text{det}} + \lambda_{\text{recon}} \cdot \mathcal{L}_{\text{recon}},
\end{equation}
where the detection loss $\mathcal{L}_{\text{det}}$ remains consistent with the previous stages, while $\mathcal{L}_{\text{recon}}$ is implemented as a Mean Squared Error (MSE) loss that quantifies the feature reconstruction quality after compression. The compression network adopts a UNet architecture to maintain spatial feature relationships during compression and decompression. Herein, we set $\lambda_{\text{recon}}=1.0$.

For network optimization, the Adam optimizer \cite{adam} is adopted across all three stages. In the first stage, the initial learning rate is set to 0.002 with 10$\times$ decay at epochs 15 and 30, and the models are trained for 40 epochs. The second and third stages both employ a fixed learning rate of 0.001 for 10 epochs. A batch size of 2 is used throughout the entire training process. This sequential training approach allows each component to specialize in its intended functionality while maintaining overall system coherence.

\section{Visualization Results}
Notably, in all visualization results below, red boxes represent prediction results, and green boxes represent ground truth.
\subsection{Qualitative Comparison Under Asynchronous Conditions}
\subsubsection{Turning Scenario}
\begin{figure*}[!t]
    \centering
    \includegraphics[width=1\textwidth]{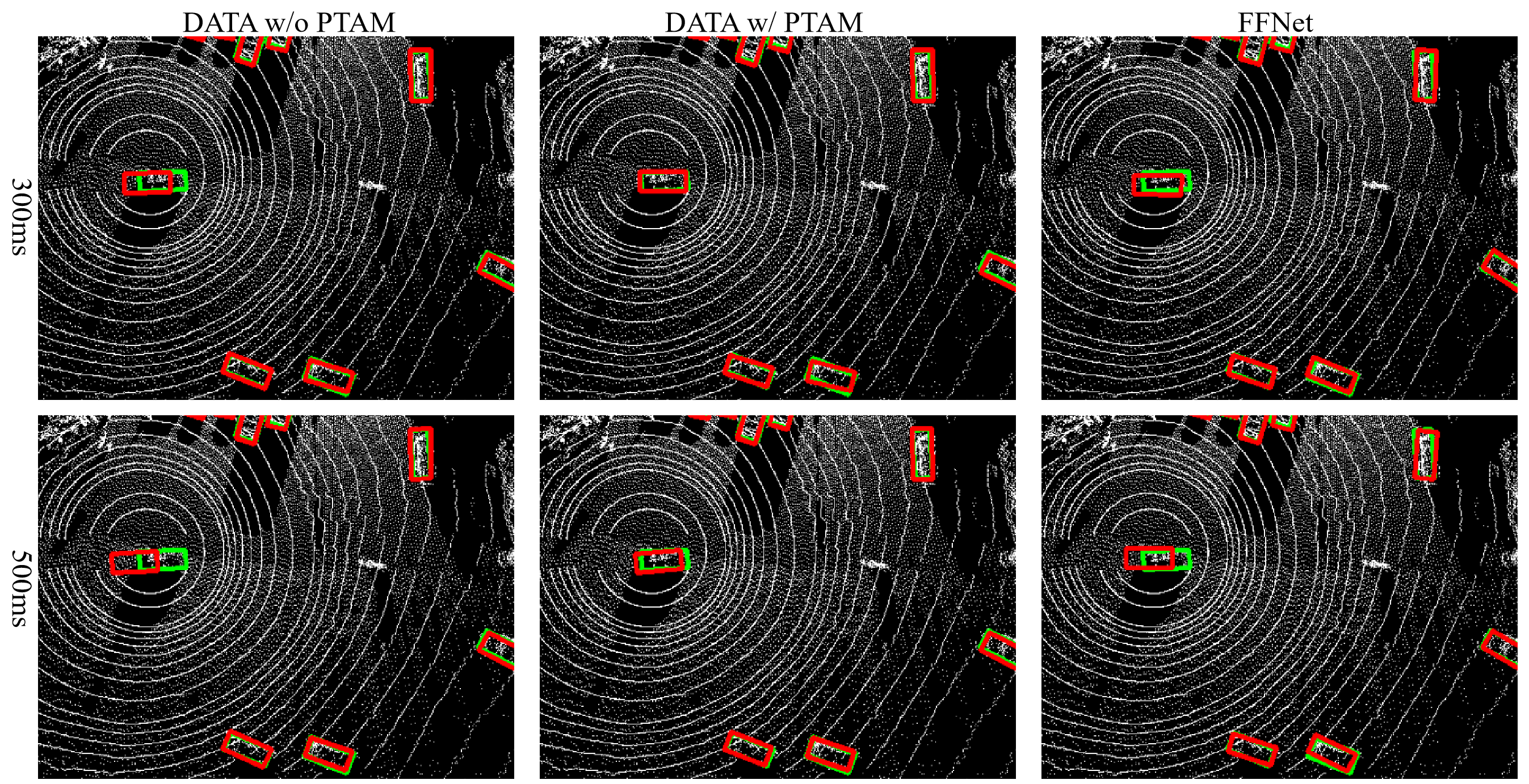}
    \vspace{-0.8cm}
    \caption{Visualization of different methods under various latency (Scene 1: Turning). This figure illustrates the collaborative process between a vehicle and the road infrastructure in a turning scenario. DATA showcases impressive ability to model variable-speed movements evidenced by precise compensation under severe latency. A significant improvement can be observed by adding PTAM to align the temporal desynchrony.}
    \label{time_001003_anno}
\end{figure*}

Figure \ref{time_001003_anno} illustrates the delay compensation performance of DATA and FFNet in turning scenarios at 300~ms and 500~ms delays. Since turning generally involves variable-speed motion, there are higher requirements for the network to model complex movements. In Figure \ref{time_001003_anno}, it shows that DATA exhibits good prediction performance under these severely delayed conditions, while the prediction results of FFNet demonstrate limited capability to accurately compensate the mismatches under such severely delayed conditions. Also, the ablation results of PTAM show significant performance improvements of detection when using PTAM, demonstrating the robustness of PTAM to acquire high-quality features of temporal alignment.
\subsubsection{Intersection Scenario}
\begin{figure*}[!t]
    \centering
    \includegraphics[width=1\textwidth]{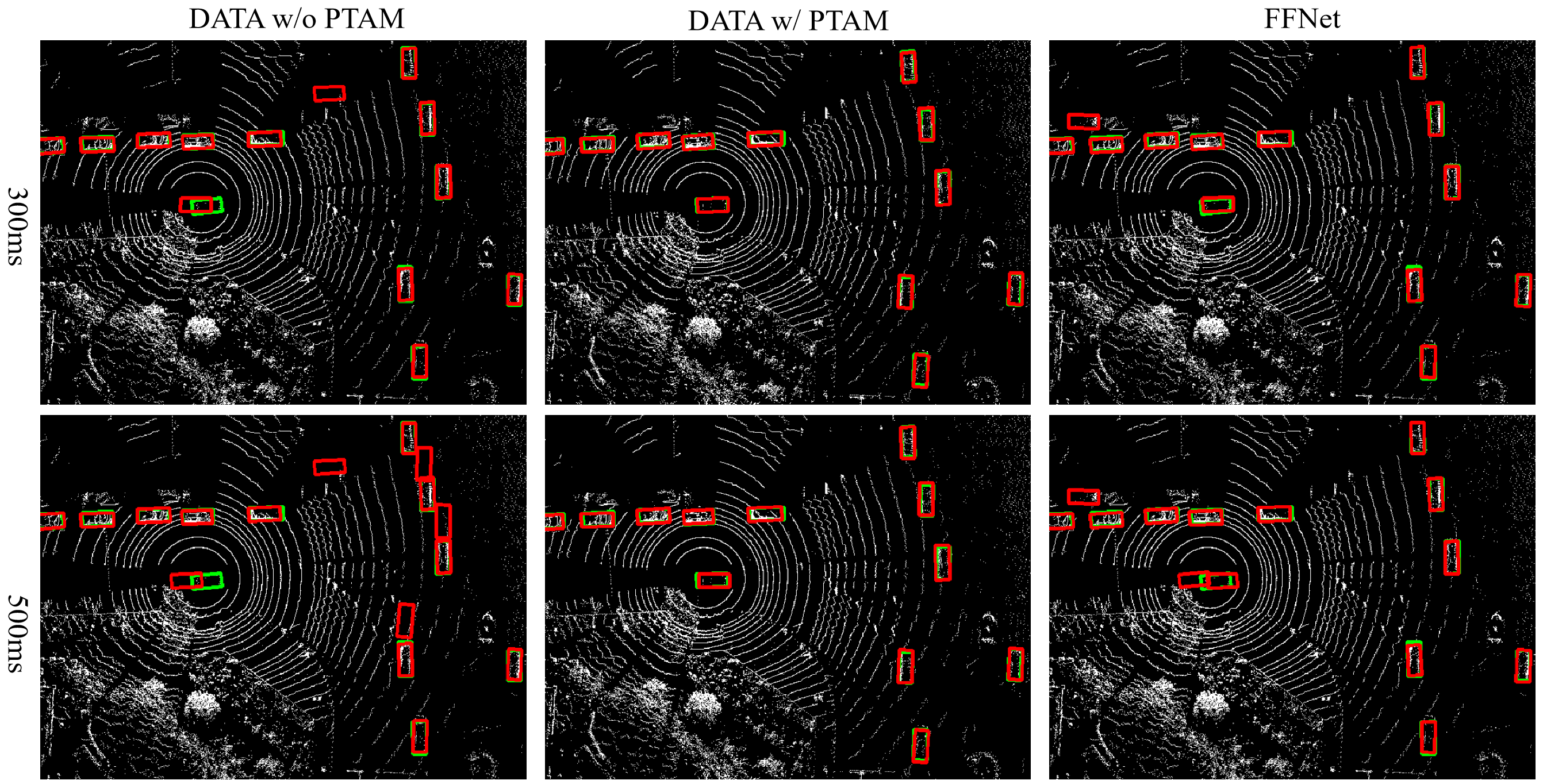}
    \vspace{-0.8cm}
    \caption{Visualization of different methods under various latency (Scene 2: Intersection). This figure depicts a scene where traffic flows intersect, challenging models to simultaneously capture motion trends in different regions. The result showcases that DATA aligns motions of two distinct directions properly, proving reliable capabilities in processing motion information of a scene-wide region.}
    \label{time_010873_anno}
\end{figure*}

The figure \ref{time_010873_anno} further exhibits the performance of DATA and FFNet under conditions with intersecting traffic flows, and this requires models to simultaneously capture motion trends in different regions. As observed from right bottom subgraph of Figure \ref{time_010873_anno}, the misalignment caused by latency can reach one vehicle length, posing significant challenges to reliable perception in autonomous driving. After implementing PTAM, DATA successfully achieves displacement compensation in both directions (horizontal and vertical directions in the figure), exhibiting both local and global motion modeling capability. In this scenario, FFNet effectively compensates for the traffic flow in the vertical direction but performs inadequately in compensation for horizontal displacement, reflecting potential deficiencies in local motion modeling that may result from its global supervision approach.

\subsection{Qualitative Comparison Under Synchronous Conditions}

\begin{figure*}[!t]
    \centering
    \includegraphics[width=1\textwidth]{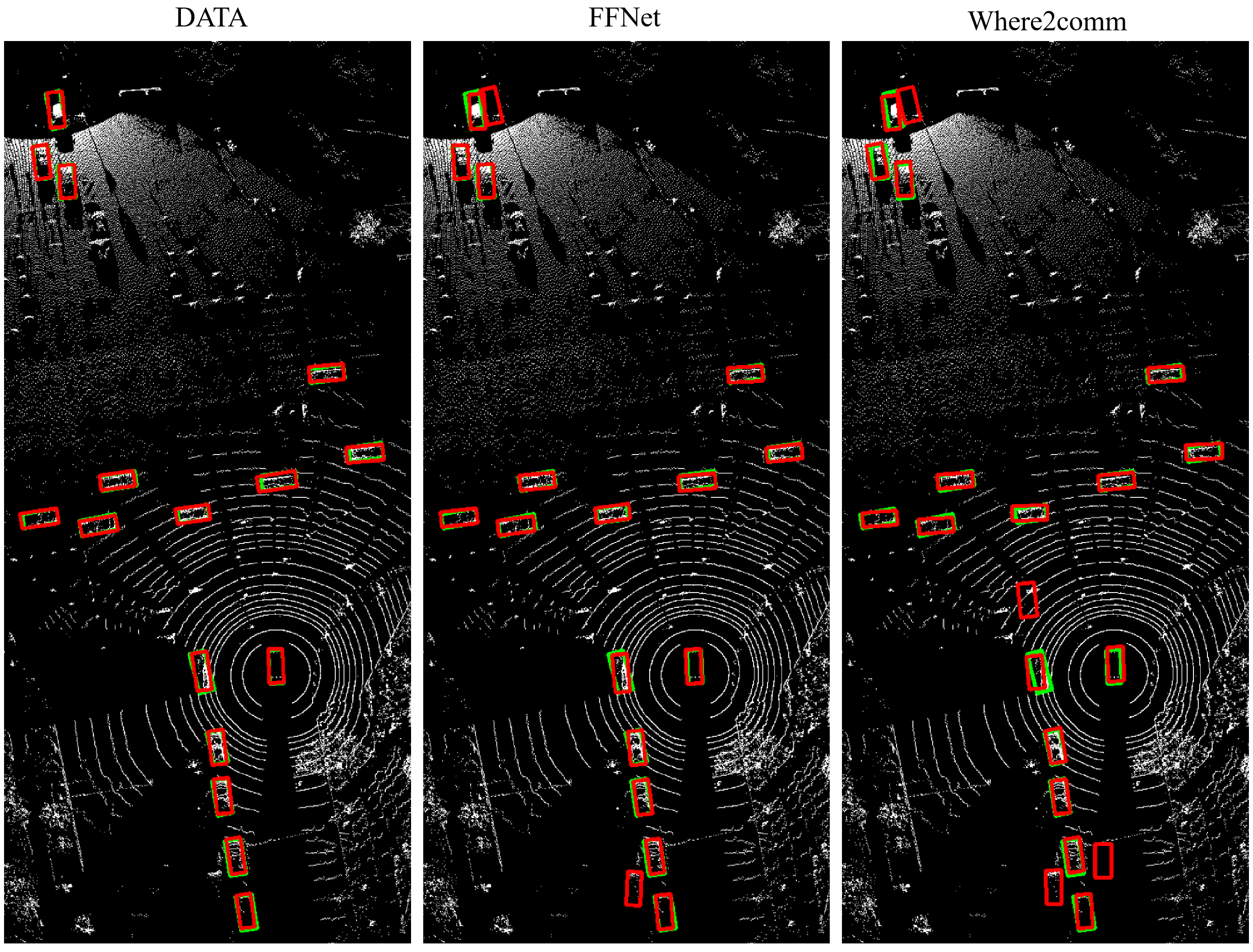}
    \vspace{-0.8cm}
    \caption{Visualization of different methods on DAIR-V2X. A vehicle-infrastructure collaborative perception scenario from DAIR-V2X shows that DATA outperforms FFNet and Where2comm by maintaining accurate detection across all regions despite domain gaps, demonstrating superior domain-invariant feature extraction under mixed observability challenges.}
    \label{dairv2x_supp}
    \vspace{-0.6cm}
\end{figure*}

Figure \ref{dairv2x_supp} illustrates a scenario from the DAIR-V2X dataset where object groups are positioned in three distinct regions: near the roadside infrastructure (collaborator), around the ego vehicle, and in the intermediate zone between them.  This configuration creates three distinct perceptual regions: areas predominantly observed by the ego vehicle, areas mainly observed by the collaborator, and overlapping areas observed by both agents. This mixed observability pattern presents a significant challenge to the perception system's ability to extract domain-invariant features for robust perception. DATA, FFNet, and Where2comm all demonstrate effective detection in areas where both the point clouds of ego and collaborator exist. However, FFNet and Where2comm generate false positives in regions dominated by point clouds either from  ego egent or collaborative agent, while DATA maintains accurate detection performance. This demonstrates that DATA effectively learns domain-invariant features through density-consistent and observability-consistent domain alignment, exhibiting stable detection performance in scenarios with significant domain gaps.

\begin{figure*}[!t]
    \centering
    \includegraphics[width=1\textwidth]{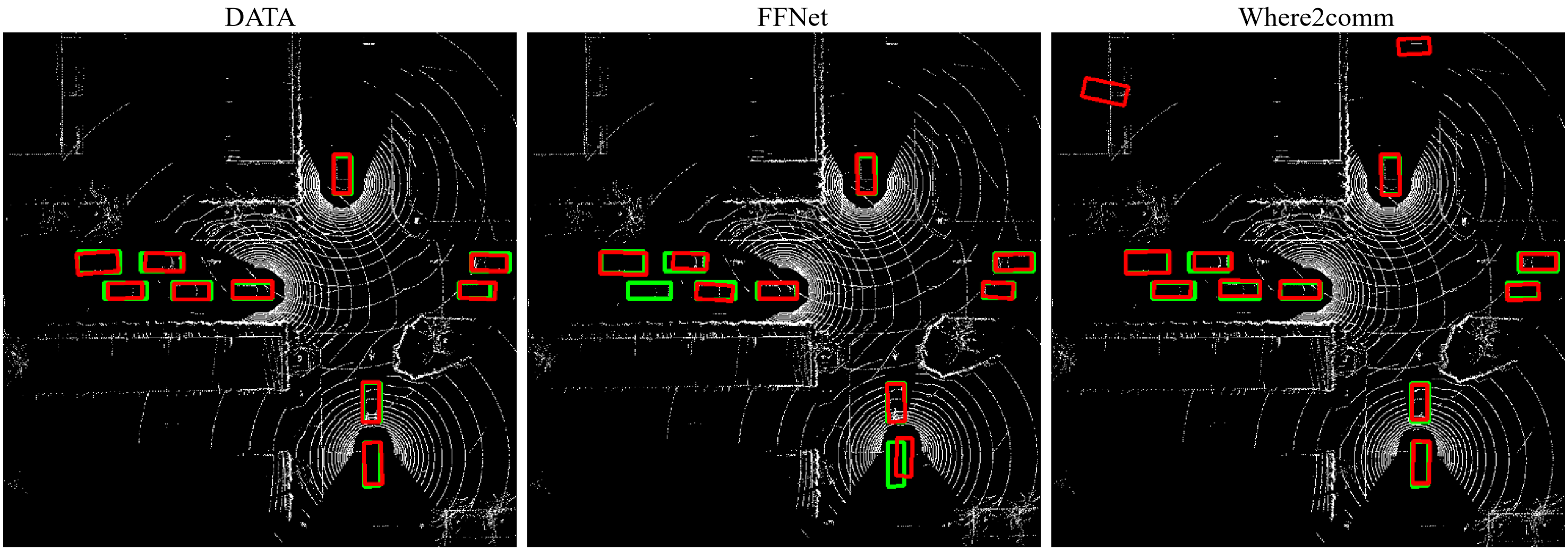}
    \vspace{-0.8cm}
    \caption{Visualization of different methods on V2XSIM. In a complex V2XSIM environment with two vehicles and infrastructure collaboration, DATA delivers superior perception by avoiding the missed detections and false positives that plague competitors when handling occluded objects and sparse point clouds.}
    \label{v2xsim_supp}
    \vspace{-0.6cm}
\end{figure*}

\begin{figure*}[!t]
    \centering
    \includegraphics[width=1\textwidth]{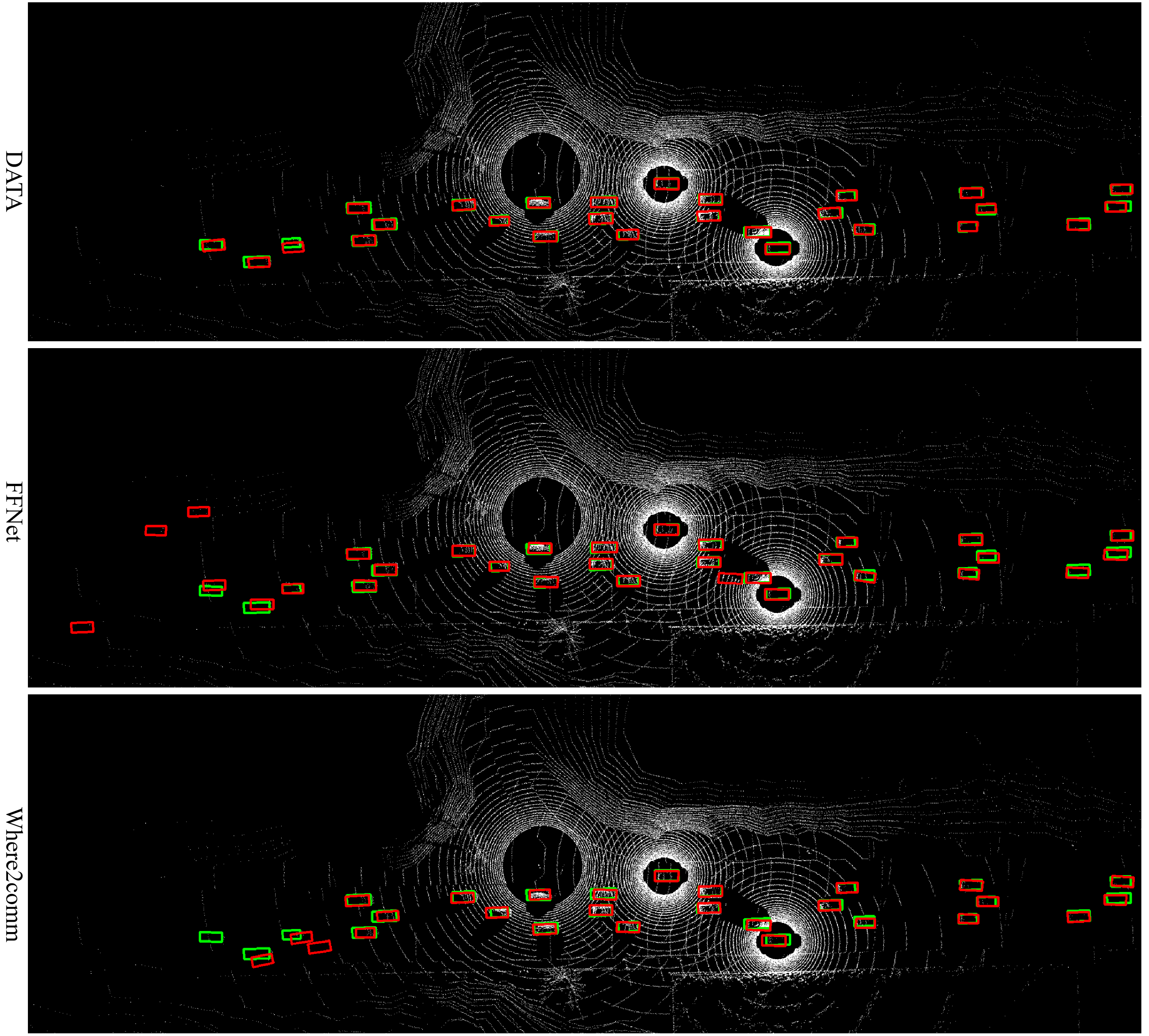}
    \vspace{-0.8cm}
    \caption{Visualization of different methods on V2XSET. In a demanding V2XSET scenario with two vehicles and roadside infrastructure collaborating, DATA outperforms competitors by accurately detecting objects despite dense traffic, occlusions, and sparse distant point clouds, demonstrating superior noise mitigation and domain-invariant feature learning.}
    \label{v2xset_supp}
    \vspace{-0.6cm}
\end{figure*}

Figure \ref{v2xset_supp} presents a demanding collaborative perception scenario with a $280~m$ range and noise interference from the V2XSET dataset. The scene contains dense traffic conditions with numerous occlusions and point cloud sparsity at extended distances, creating an ideal testing environment for evaluating collaborative perception capabilities. FFNet and Where2comm achieve good perception results in regions where collaborative agents provide reliable information, but exhibit detection inaccuracies and false positives at greater distances due to point cloud sparsity and noise interference. DATA effectively mitigates noise interference and the challenges of sparse long-distance point clouds through learned domain-invariant features and foreground enhancement, achieving robust detection performance across the entire extended range.
Figure \ref{v2xsim_supp} similarly demonstrates a noisy scenario from V2XSIM with occlusions and sparse point clouds. DATA achieves effective perception performance, while FFNet and Where2comm exhibit missed detections and false positives in regions with sparse point clouds, further validating DATA's robust and reliable perception capabilities.

\begin{table*}[!t]
\centering
\begin{tabular}{|p{2.8cm}|p{7.2cm}|p{2.5cm}|}
\hline
\textbf{Symbol} & \textbf{Definition} & \textbf{Section/Module} \\
\hline
\hline

\multicolumn{3}{|l|}{\textbf{3.1 Problem Formulation and Overall Architecture}} \\
\hline
$N$ & Number of agents in collaborative perception system & Problem Formulation \\
\hline
$i, j$ & Subscripts denoting ego agent and collaborative agents, where $i \neq j$ & Problem Formulation \\
\hline
$X_i(t)$ & Latest data of ego agent at current time $t$ & Problem Formulation \\
\hline
$X_j(t-\tau)$ & Data transmitted by collaborative agents at timestamp $t-\tau$ & Problem Formulation \\
\hline
$X_j(t-\tau-\Delta T)$ & Data of collaborative agent at timestamp $t-\tau-\Delta T$ & Problem Formulation \\
\hline
$\tau$ & Transmission delay & Problem Formulation \\
\hline
$\Delta T$ & Time interval & Problem Formulation \\
\hline
\hline

\multicolumn{3}{|l|}{\textbf{3.2.1 Proximal-region Hierarchical Downsampling (PHD)}} \\
\hline
$\mathcal{O}_i$ & Set of all objects observable to ego agent $i$ & CDAM-PHD \\
\hline
$N^{total}_i$ & Total number of observable objects & CDAM-PHD \\
\hline
$o_k$ & $k$-th object & CDAM-PHD \\
\hline
$\mathcal{O}^{prox}_i$ & Objects within the proximal region & CDAM-PHD \\
\hline
$d_i(o_k)$ & Distance from ego agent $i$ to $k$-th object & CDAM-PHD \\
\hline
$d_{th}$ & Distance threshold & CDAM-PHD \\
\hline
$N_{proc}$ & Number of objects selected from $\mathcal{O}^{prox}_i$ for subsequent processing & CDAM-PHD \\
\hline
$N_{max}$ & Predefined maximum number of objects & CDAM-PHD \\
\hline
$B^{out}_{i,k}$ & Oriented bounding box of $k$-th object & CDAM-PHD \\
\hline
$B^{in}_{i,k}$ & Inner bounding box with scaling factor $\alpha$ & CDAM-PHD \\
\hline
$(x_k, y_k, z_k)$ & Center coordinates of $k$-th object & CDAM-PHD \\
\hline
$(h_k, w_k, l_k)$ & Dimensions (height, width, length) of $k$-th object & CDAM-PHD \\
\hline
$\theta_k$ & Orientation of $k$-th object & CDAM-PHD \\
\hline
$\alpha$ & Scaling factor to adjust bounding box dimensions, $\alpha \in (0,1)$ & CDAM-PHD \\
\hline
$R^{in}_{i,k}, R^{out}_{i,k}$ & Point sets in inner and outer regions & CDAM-PHD \\
\hline
$\beta_{in}, \beta_{out}$ & High and conservative downsampling ratios for inner and outer regions & CDAM-PHD \\
\hline
$\text{FPS}(\cdot,\cdot)$ & Farthest Point Sampling operation & CDAM-PHD \\
\hline
$\tilde{R}^{in}_{i,k}, \tilde{R}^{out}_{i,k}$ & Downsampled inner and outer regions & CDAM-PHD \\
\hline
$\tilde{P}_{i,k}$ & Point clouds of $k$-th object after hierarchical downsampling & CDAM-PHD \\
\hline

\end{tabular}
\caption{Notation Table for DATA Paper - Part I: Problem Formulation and PHD}
\label{tab:notation1}
\end{table*}

\begin{table*}[!t]
\centering
\begin{tabular}{|p{2.3cm}|p{7.7cm}|p{2.5cm}|}
\hline
\textbf{Symbol} & \textbf{Definition} & \textbf{Section/Module} \\
\hline
\hline

\multicolumn{3}{|l|}{\textbf{3.2.2 Observability-constrained Discriminator (OD)}} \\
\hline
$H_i, H_j$ & BEV features of ego and collaborative agents & CDAM-OD \\
\hline
$C, H, W$ & Channel dimension, height, and width of features & CDAM-OD \\
\hline
$M_i, M_j$ & Observability maps, $M_i, M_j \in \mathbb{R}^{1 \times H \times W}$ & CDAM-OD \\
\hline
$\Phi(\cdot)$ & Foreground estimator & CDAM-OD \\
\hline
$H_{j \rightarrow i}, M_{j \rightarrow i}$ & Collaborative features and observability map projected onto ego coordinates & CDAM-OD \\
\hline
$\mathcal{V}$ & Set of valid grids in transformed feature map & CDAM-OD \\
\hline
$H^{comp}_j, M^{comp}_j$ & Complemented feature and observability map & CDAM-OD \\
\hline
$I_{\mathcal{V}}$ & Indicator function (equals 1 for points in $\mathcal{V}$ and 0 for others) & CDAM-OD \\
\hline
$W$ & Observability weighting map, $W \in \mathbb{R}^{1 \times H \times W}$ & CDAM-OD \\
\hline
$\mathcal{L}_{domain}$ & Domain alignment objective & CDAM-OD \\
\hline
$\mathcal{S}$ & Set of all spatial positions & CDAM-OD \\
\hline
$sp$ & One spatial position & CDAM-OD \\
\hline
$W_{flat}$ & Flattened observability weighting map & CDAM-OD \\
\hline
$\Psi_\theta$ & Feature extractor (point clouds as input and BEV features as output) & CDAM-OD \\
\hline
$D_\mu$ & Discriminator & CDAM-OD \\
\hline
$\mathcal{L}_{BCE}$ & Binary cross-entropy loss & CDAM-OD \\
\hline
$Z$ & Domain label (0 for ego agent and 1 for collaborative agent) & CDAM-OD \\
\hline
$\gamma$ & Negative scaling factor in gradient reversal layer ($\gamma = -0.1$) & CDAM-OD \\
\hline

\end{tabular}
\caption{Notation Table for DATA Paper - Part II: Observability-constrained Discriminator}
\label{tab:notation2}
\end{table*}

\begin{table*}[!t]
\centering
\begin{tabular}{|p{2.8cm}|p{7.2cm}|p{2.5cm}|}
\hline
\textbf{Symbol} & \textbf{Definition} & \textbf{Section/Module} \\
\hline
\hline

\multicolumn{3}{|l|}{\textbf{3.3 Progressive Temporal Alignment Module (PTAM)}} \\
\hline
$F_j(t, x)$ & Features at collaborative agent at time $t$ and position $x$ & PTAM \\
\hline
$v(t-\Delta t, x)$ & Velocity field describing motion of features at time $t-\Delta t$ & PTAM \\
\hline
$\Delta t$ & Time interval within range of typical transmission delay & PTAM \\
\hline
$\Delta p$ & Motion field representing velocity, $\Delta p \in \mathbb{R}^{2 \times H \times W}$ & PTAM \\
\hline
$\xi$ & Temporal scaling factor corresponding to $\Delta t$ & PTAM \\
\hline
$F^{inter}_j$ & Intermediate feature predicted by collaborative agent & PTAM \\
\hline
$\hat{F}_j(t-\tau)$ & Feature from collaborative agent $j$ at time $t-\tau$, as received by ego agent & PTAM \\
\hline
$\hat{F}^{inter}_j$ & Intermediate feature from collaborative agent $j$, as received by ego agent & PTAM \\
\hline
$\hat{F}_j(t)$ & Temporally aligned features & PTAM \\
\hline
$F_{latest}, F_{prev}$ & Latest feature and its previous feature & PTAM \\
\hline
$\Delta F$ & Temporal feature difference, $\Delta F = F_{latest} - F_{prev}$ & PTAM \\
\hline
$w_{samp}$ & Sampling weight, $w_{samp} \in \mathbb{R}^{1 \times H \times W}$ & PTAM \\
\hline
$f_{warp}(\cdot)$ & Bilinear sampling operation & PTAM \\
\hline
$s1, s2$ & Superscripts indicating first stage and second stage & PTAM \\
\hline
$\Delta M$ & Motion difference field & PTAM \\
\hline
$f_M$ & Global context vector & PTAM \\
\hline
$f_T$ & Temporal encoding with sinusoidal positional embeddings & PTAM \\
\hline
\hline

\multicolumn{3}{|l|}{\textbf{3.3.3 Multi-window Self-supervised Training Strategy}} \\
\hline
$s$ & Spatial scale index & PTAM Training \\
\hline
$h_s, w_s$ & Feature plane size at spatial scale $s$ & PTAM Training \\
\hline
$l$ & Window size & PTAM Training \\
\hline
$\mathcal{W}_1, \mathcal{W}_2$ & Two complementary window partitioning strategies & PTAM Training \\
\hline
$w_{m',n'}$ & Window with top-left corner at position $(m', n')$ & PTAM Training \\
\hline
$w_{p',q'}$ & Window with top-left corner at position $(p', q')$ & PTAM Training \\
\hline
$N_{window}$ & Total number of windows & PTAM Training \\
\hline
$F^{gt}_j$ & Ground truth features & PTAM Training \\
\hline
$\cos(\cdot,\cdot)$ & Cosine similarity between predictions and ground truth features & PTAM Training \\
\hline
$\mathcal{L}^s_{inter}, \mathcal{L}^s_{final}$ & Loss functions for intermediate and final predictions at scale $s$ & PTAM Training \\
\hline
$\mathcal{L}_{temporal}$ & Total temporal alignment loss computed across all three scales & PTAM Training \\
\hline

\end{tabular}
\caption{Notation Table for DATA Paper - Part III: Progressive Temporal Alignment Module}
\label{tab:notation3}
\end{table*}

\begin{table*}[!t]
\centering
\begin{tabular}{|p{2.3cm}|p{7.7cm}|p{2.5cm}|}
\hline
\textbf{Symbol} & \textbf{Definition} & \textbf{Section/Module} \\
\hline
\hline

\multicolumn{3}{|l|}{\textbf{3.4 Instance-focused Feature Aggregation Module (IFAM)}} \\
\hline
$H_a$ & BEV feature of $a$-th agent, $H_a \in \mathbb{R}^{C \times H \times W}$ & IFAM \\
\hline
$M_a$ & Foreground mask, $M_a = \Phi(H_a)$ & IFAM \\
\hline
$H^{fore}_a, H^{back}_a$ & Foreground and background features & IFAM \\
\hline
$H^{enh}_a$ & Enhanced foreground features after structural convolution & IFAM \\
\hline
$\text{StructConv}(\cdot)$ & Structural convolution with specialized convolutions & IFAM \\
\hline
$H^{cat}_a$ & Concatenated features, $H^{cat}_a \in \mathbb{R}^{2C \times H \times W}$ & IFAM \\
\hline
$W^s_a, W^c_a$ & Spatial attention and channel attention & IFAM \\
\hline
$W^{init}_a$ & Initial attention weights, $W^{init}_a = W^s_a \oplus W^c_a$ & IFAM \\
\hline
$W^{verif}_a$ & Verification weights & IFAM \\
\hline
$\text{CS}$ & Channel shuffle operation & IFAM \\
\hline
$\text{GConv}(\cdot)$ & Group convolution for independent weight generation & IFAM \\
\hline
$H^{verif}_a$ & Verified foreground feature & IFAM \\
\hline
$H^{refined}_a$ & Individually refined BEV features & IFAM \\
\hline
$\epsilon$ & Learnable parameter balancing contribution of background features & IFAM \\
\hline
\hline

\multicolumn{3}{|l|}{\textbf{4. Experiments and General Symbols}} \\
\hline
$\text{AP}_{50}, \text{AP}_{70}$ & Average Precision at IoU thresholds of 0.50 and 0.70 & Evaluation Metrics \\
\hline
$\text{IoU}$ & Intersection-over-Union & Evaluation Metrics \\
\hline
$\mu_{pos}, \sigma_{pos}$ & Mean and standard deviation for position noise & Noise Modeling \\
\hline
$\mu_{rot}, \sigma_{rot}$ & Mean and standard deviation for orientation noise & Noise Modeling \\
\hline
$\sigma_{local}$ & Standard deviation for localization noise & Robustness Testing \\
\hline
$\sigma_{head}$ & Standard deviation for heading noise & Robustness Testing \\
\hline
$\odot$ & Element-wise product & General Operations \\
\hline
$\oplus$ & Concatenation operation & General Operations \\
\hline
$[\cdot]$ & Concatenation along specified dimension & General Operations \\
\hline
$|\cdot|$ & Cardinality of a set & General Operations \\
\hline
$\text{softmax}(\cdot)$ & Softmax function & Activation Functions \\
\hline
$\text{ReLU}(\cdot)$ & ReLU activation function & Activation Functions \\
\hline
$\text{Sigmoid}(\cdot)$ & Sigmoid function & Activation Functions \\
\hline
$\text{MLP}(\cdot)$ & Multi-layer perceptron & Network Structures \\
\hline
$\text{Conv}_{1 \times 1}(\cdot)$ & $1 \times 1$ convolution operation & Network Structures \\
\hline

\end{tabular}
\caption{Notation Table for DATA Paper - Part IV: IFAM and General Symbols}
\label{tab:notation4}
\end{table*}

\end{document}